\documentclass{article} % For LaTeX2e
\usepackage{iclr2026_conference,times}

\usepackage{amsmath,amsfonts,bm}

\def\eqref#1{equation~\ref{#1}}
\def\1{\bm{1}}

\DeclareMathAlphabet{\mathsfit}{\encodingdefault}{\sfdefault}{m}{sl}
\SetMathAlphabet{\mathsfit}{bold}{\encodingdefault}{\sfdefault}{bx}{n}

\usepackage[table]{xcolor}
\usepackage{hyperref}
\usepackage{cleveref}
\usepackage{tocloft}
\usepackage{url}
\usepackage{multirow}
\usepackage{tcolorbox}
\usepackage{mathtools}
\usepackage{booktabs}
\usepackage{tabularx}
\usepackage{array}
\usepackage{tikz}
\usepackage{caption}
\usepackage{graphicx}
\usepackage{subcaption}
\usepackage{listings}
\usepackage{xcolor}
\usepackage{algorithm}
\usepackage{algorithmicx}
\usepackage{wrapfig}
\usepackage{pifont}

\definecolor{mygreen}{rgb}{0.0, 0.5, 0.0}
\newcommand{\cmark}{\textcolor{mygreen}{\ding{51}}}
\newcommand{\xmark}{\textcolor{red}{\ding{55}}}

\usepackage{algpseudocode}
\hypersetup{
    colorlinks=true,
    linkcolor=blue!50,
    urlcolor=blue!50,
    citecolor=blue!50
}

\lstdefinestyle{mypython}{
  language=Python,
  basicstyle=\ttfamily\scriptsize,
  backgroundcolor=\color{gray!5},
  keywordstyle=\color{blue},
  stringstyle=\color{green!60!black},
  commentstyle=\color{purple!80},
  numbers=left,
  numberstyle={\tiny\color{gray!90}},
  stepnumber=1,
  numbersep=8pt,
  frame=single,
  breaklines=true,
  tabsize=2,
  showstringspaces=false
}

\tcbset{
  promptstyle/.style={
    colback=brown!10,
    colframe=brown!50!black,
    fonttitle=\bfseries,
    coltitle=white,
    colbacktitle=brown!50!black,
    rounded corners,
    boxrule=0.75mm,
    coltext=black,
    width=\columnwidth,
    fontupper=\small
  }
}
\tcbset{
  casestyle/.style={
    colback=purple!10,
    colframe=purple!50!black,
    fonttitle=\bfseries,
    coltitle=white,
    colbacktitle=purple!50!black,
    rounded corners,
    boxrule=0.75mm,
    coltext=black,
    width=\columnwidth,
    fontupper=\small
  }
}

\title{Self-Aligned Reward: Towards Effective and Efficient Reasoners}

\author{
Peixuan Han$^{1}$\thanks{Work done during an internship at Amazon Web Services.}  \quad
Adit Krishnan$^{2}$ \quad
Gerald Friedland$^{2}$ \quad
Jiaxuan You$^{1}$ \quad
Chris Kong$^{2}$\thanks{Corresponding author.} \\
$^{1}$University of Illinois Urbana-Champaign \quad $^{2}$Amazon Web Services\\
\texttt{ph16@illinois.edu}, \texttt{luyankon@amazon.com}\\
}

\NewDocumentCommand{\peixuan}
{ mO{} }{\textcolor{purple}{\textsuperscript{\textit{peixuan}}\textsf{\textbf{\small[#1]}}}}

\NewDocumentCommand{\adit}
{ mO{} }{\textcolor{blue}{\textsuperscript{\textit{adit}}\textsf{\textbf{\small[#1]}}}}

\NewDocumentCommand{\chris}
{ mO{} }{\textcolor{orange}{\textsuperscript{\textit{chris}}\textsf{\textbf{\small[#1]}}}}

\iclrfinalcopy
\begin{document}

\maketitle

\begin{abstract}

% \peixuan{When describing percentages (e.g. 4\%), make sure it's clear comparing to GRPO instead of base}
Reinforcement learning with verifiable rewards has significantly advanced reasoning with large language models (LLMs) in domains such as mathematics and logic. However, verifiable signals provide only coarse-grained or binary correctness feedback. This limitation results in inefficiencies like overly verbose or repetitive reasoning. Existing length-based solutions (e.g., length penalty) compromise accuracy. To address this deficiency, we introduce \textbf{self-aligned reward (SAR)}, a generic, universally applicable self-guided signal that complements verifiable rewards to enhance both reasoning accuracy and efficiency in RL. Specifically, SAR is defined as the relative perplexity difference between an answer conditioned on the query and the standalone answer, thereby favoring responses that are concise and query-specific. Quantitative analysis reveals that SAR reliably judges answer quality: concise, correct answers score higher than redundant ones, and partially correct answers score higher than entirely incorrect ones. Evaluation on 4 different models across 7 benchmarks shows that integrating SAR with prevalent RL algorithms like PPO and GRPO reduces answer length by 30\%, while improving accuracy by 4\%. Our analysis also shows that SAR generalizes well to out-of-domain tasks and achieves a Pareto-optimal frontier between correctness and efficiency compared to state-of-the-art baselines. We also show that SAR shortens unnecessary elaboration while preserving advanced reasoning behaviors. These results highlight the promise of self-aligned reward as a fine-grained complement to verifiable rewards, paving the way for efficient and effective LLM training. All of our code implementations and data are publicly available at GitHub~\footnote {\url{https://github.com/amazon-science/Self-Aligned-Reward-Towards_Effective_and_Efficient_Reasoners}}.

\definecolor{mygray}{rgb}{0, 0.23, 0.35} % values between 0 and 1

\begin{figure}[htbp]
    \centering
    \includegraphics[width=\textwidth]{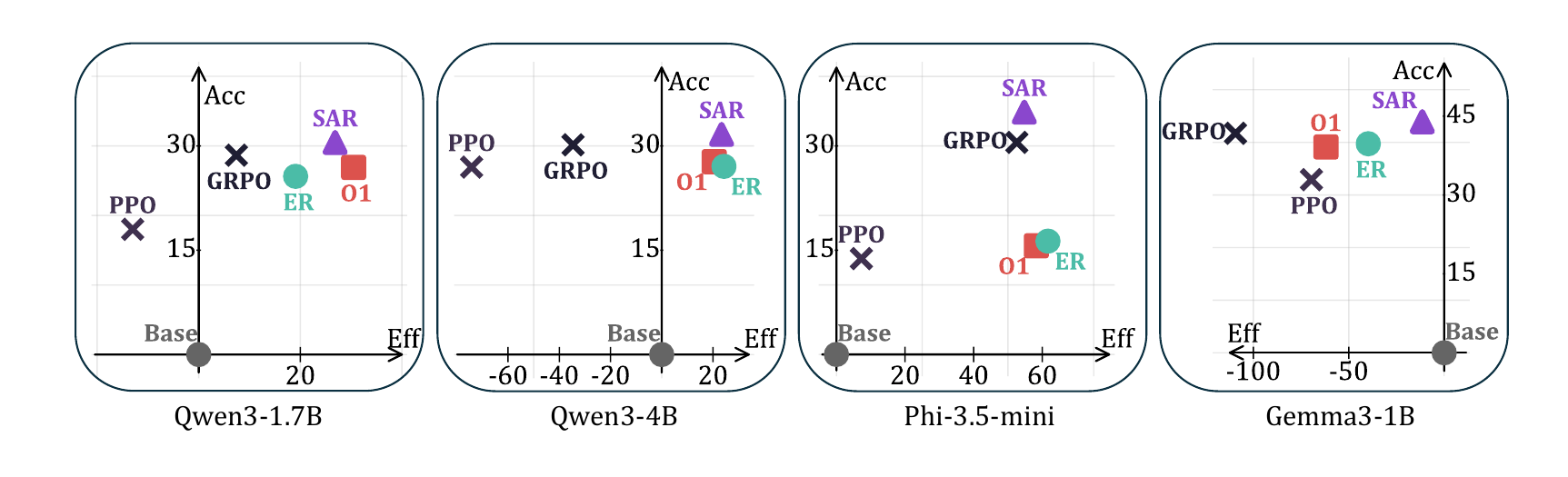}
    \vspace{-2em}
    \caption{\textbf{Training with self-aligned reward enhances both efficiency and accuracy}. We present the relative gains in efficiency and accuracy compared to the respective base model in math reasoning benchmarks. Efficiency gain is measured as the drop in average response length.}
    \label{fig:intro_fig}
    \vspace{-2em}
\end{figure}

\end{abstract}
\section{Introduction}

% {
% Q:
% A: 1 + 1 = 2
% }

% 5 bullets for structuring instruction

% \textbullet \hspace{1pt} What is the problem?

% Enhancing RL efficiency with internal signals

% \textbullet \hspace{1pt} Why is it interesting and important?

% Avoiding overthinking means less latency and cost

% Internal signals mean the method is more scalable and unbiased (self-evolving)

% Efficient reasoning has been widely discussed since deepseek-r1

% \textbullet \hspace{1pt} Why is it hard? (E.g., why do naive approaches fail?)

% Making answers shorter often means sacrificing accuracy

% Signals that don't rely on an external model is limited

% \textbullet \hspace{1pt} Why hasn’t it been solved before?

% Length-based methods don't consider answer quality; shorter isn't always better

% \textbullet \hspace{1pt} Why are the key components of my approach and results? 

% Method:52. SAPO, using a reward to measure relative perplexity drop between a and a|q

%  - R\_sa promotes correct and concise answers

%  - R\_sa can distinguish useful reasoning from redundant content

% SAPO shows improved efficiency and accuracy across 4 models

Recently, reinforcement learning (RL) with verifiable rewards has attracted broad attention in LLM training, showing remarkable improvements in reasoning skills~\citep{guo2025deepseek,jaech2024openai}. However, such verifiable signals are inherently discrete and coarse: they only judge final answer correctness, but fail to capture finer distinctions among responses. For instance, an unnecessarily long solution receives no penalty as long as the final answer is correct, and an almost correct response is treated the same as a completely wrong one. This limitation often induces ``overthinking'', where models generate unnecessary elaborations that increase latency and cost~\citep{sui2025stop}. 

To this end, researchers have proposed heuristic regulations such as length penalties or brevity-oriented objectives~\citep{luo2025o1,aggarwal2025l1}. While effective in reducing output verbosity, these methods often penalize both redundant and essential reasoning, thereby harming accuracy when necessary intermediate steps are suppressed. Consequently, this line of approaches struggles to balance efficiency with correctness. Using external signal sources, such as reward models, is also undesirable due to their vulnerability to reward hacking. This underscores the necessity of developing internally grounded reward mechanisms that provide precise and detailed guidance, discerning necessary reasoning from redundant elaboration.

% \begin{figure}[htbp]
% \vspace{-10pt}
%     \centering
%     \includegraphics[width=0.76\textwidth]{figures/Intro-Img.pdf}
% \vspace{-5pt}
%         \caption{The above figure illustrates the key differences between Proximal Policy Optimization (PPO), Group Relative Policy Optimization (GRPO) with Verifiable-Rewards, and our proposal for Self-Aligned Policy Optimization (SAPO).}
%         \label{fig:intro}
% \end{figure}

\begin{table}[t]
\caption{Comparison of different reward designs.}
\vspace{-0.4em}
\label{table:compare}
\small
\centering
\renewcommand{\arraystretch}{1.05}
\setlength{\tabcolsep}{4pt} % Reduces space to 3pt

\resizebox{1.0\textwidth}{!}{%
\begin{tabular}{lcccccc}
\toprule
Reward               & Continuous & Internal & Content-Aware & Correctness & Conciseness \\\hline
Correctness             &   \xmark           &    \cmark       &      \cmark          &               \cmark      &       \xmark              \\
Reward Model \citep{ouyang2022training}              &     \cmark        &    \xmark        &          \cmark      &         \cmark            &        \xmark             \\
Length Penalty \citep{chen2025overthinker}           &     \cmark         &    \cmark       &         \xmark       &         \xmark            &     \cmark                \\
Entropy \citep{agarwal2025unreasonable}       &      \cmark        &     \cmark      &        \cmark        &          \xmark           &      \xmark               \\
\textbf{Self-Aligned (Ours)}            &   \cmark           &   \cmark        &        \cmark        &               \cmark      &     \cmark               \\\bottomrule
\end{tabular} }

\end{table}

% Of the various internal signals, \textit{perplexity} offers a promising option, given its role as a natural proxy for model confidence~\citep{friedland2024pplqa,agarwal2025unreasonable}. Building on this insight,

To close this gap, we introduce \textbf{Self-Aligned Reward (SAR)}, a self-guided proxy to judge answer quality (\Cref{eq:pplqa}) based on \textit{perplexity}, an informative metric modeling uncertainty~\citep{friedland2024pplqa,agarwal2025unreasonable}. Specifically, SAR evaluates the perplexity of an answer both in isolation and when conditioned on the query, and then measures their relative difference between the two. Consequently, the reward promotes answers that are highly confident under the query context but unlikely to arise independently without the query, which typically corresponds to responses that are concise and strongly aligned with the query. Notably, SAR is the only fine-grained approach that promotes accuracy and efficiency at the same time, as shown in \Cref{table:compare}.

We first conduct a quantitative analysis of different types of answers to demonstrate that SAR provides an accurate fine-grained reward landscape over answers of different qualities (\Cref{sec:case_study}). We then train LLMs by combining SAR and verifiable reward in PPO and GRPO, two prevalent reinforcement learning algorithms. \textbf{We find PPO and GRPO with SAR} (denoted as \textbf{SA-PPO} and \textbf{SA-GRPO}) \textbf{achieve notable gains over baselines across 4 models and 7 benchmarks}, improving accuracy by 4\% and efficiency by 30\% (\Cref{sec:exp_result}). Moreover, SAR outperforms length-based rewards with a Pareto-optimal front in the accuracy-efficiency trade-off (\Cref{exp_tradeoff}). In addition, we demonstrate the advantages of SAR over confidence-based methods (\Cref{sec:ablation}) and provide an analysis of its reasoning behaviors (\Cref{sec:reasoning}). Our findings suggest that combining verifiable rewards with intrinsic model self-judgment offers a new paradigm for RL training, enabling improvements in both reasoning capability and efficiency.

\section{Related Work}

\textbf{Reinforcement Learning for LLMs.} Reinforcement learning (RL) has emerged as a powerful paradigm for fine-tuning large language models (LLMs) to enhance their performance in reasoning tasks~\citep{ouyang2022training,guo2025deepseek}. Proximal Policy Optimization (PPO)~\citep{schulman2017proximal} and Group Relative Policy Optimization (GRPO)~\citep{shao2024deepseekmath} are typical algorithms that are widely adopted in diverse scenarios and domains~\citep{kulkarni2024reinforcement,chen2025rm,han2025tomap,liu2025time}. Improvements to these algorithms have also been proposed, namely Dr.GRPO~\citep{liu2025understanding}, GSPO~\citep{zheng2025groupsequencepolicyoptimization}, and Clip-Cov~\citep{cui2025entropy}. 

% RL has shown great potential in many fields such as math reasoning~\citep{shao2024deepseekmath}, retrieval-augmented generation~\citep{kulkarni2024reinforcement}, reward modeling~\citep{chen2025rm}, debating~\citep{han2025tomap}, and temporal reasoning~\citep{liu2025time}.

% Proximal Policy Optimization (PPO)~\citep{schulman2017proximal} is a widely adopted RL algorithm, leveraging an actor-critic framework to optimize policies while maintaining stability through clipped objectives and a Kullback-Leibler (KL) divergence penalty. However, PPO's reliance on a separate critic network increases computational and memory demands, making it resource-intensive for large-scale LLM training. 

% In addition, Simplerl-zoo systematically analyzes the impact of base model and training settings on RL~\citep{zeng2025simplerl}. 

\textbf{Efficient Reasoning.} Reasoning models often suffer from overthinking~\citep{su2025between,cuadron2025danger,sui2025stop}, leading to unnecessary computation burdens. Prompt engineering~\citep{han2024token,ma2025reasoning,renze2024benefits} and instruction tuning~\citep{yu2024distilling,kang2025c3ot,xia2025tokenskip,han2024token,ma2025cot} have been widely used to mitigate this drawback.
% The first direction is prompt engineering, where techniques such as Token-Budget~\citep{han2024token}, No-Thinking~\citep{ma2025reasoning}, and Concise CoT~\citep{renze2024benefits} are proposed to enforce concise reasoning by limiting token usage or structuring outputs. Secondly, instruction-tuning LLMs on efficient reasoning traces enables models to learn compact reasoning paths~\citep{yu2024distilling,kang2025c3ot,xia2025tokenskip}. \citet{han2024token,ma2025cot} enables users to specify expected reasoning lengths, making the model more flexible.
Recently, researchers have explored reward signals in RL to penalize lengthy answers~\citep{aggarwal2025l1,team2025kimi} or relative lengths~\citep{luo2025o1,arora2025training}. In addition, \citet{yeo2025demystifying} explores the relationship of the length penalty with reasoning behaviors and \citet{chen2025overthinker} proposes a difficulty-sensitive method for token compression. However, these methods sacrifice accuracy to gain efficiency, which is a key issue this paper aims to address.

\textbf{Self-judging of LLMs.} The self-judgment capabilities of LLMs~\citep{renze2024self} play a key role in scalable training and self-evolution. For instance, self-confidence estimation is widely used to reduce hallucination~\citep{geng2023survey,wen2024mitigating,ji2023towards}, probing internal states provides a lens to analyze model behavior~\citep{han2025safeswitch,chen2024inside}, and majority voting~\citep{wang2022self} is critical in inference-time scaling techniques~\citep{snell2024scaling}. Self-judging has also been applied to train reasoning models, with a primary focus on entropy-based metrics~\citep{agarwal2025unreasonable,lei2025revisiting,zhao2025learning}. \citet{zhang2025no} provides a theoretical foundation to self-judged signals. % Our work also focuses on this direction, and we novelly propose the self-aligned reward measuring question-answer relevancy instead of self-confidence, and combine it with verifiable signals.

%but is fundamentally different from previous approaches. We don't eliminate verifiable signals, but combine it with self-judged signals for better performance. In addition, our self-aligned reward measures ``confidence gain'' when conditioning on the question, instead of confidence itself.

% original pplqa isn't self-judge, so we can't include that here
\section{Methodology}

\subsection{Reinforcement Learning Formulation}
\label{method:rl}

Proximal Policy Optimization (PPO)~\citep{schulman2017proximal} and Group Relative Policy Optimization (GRPO)~\citep{shao2024deepseekmath} are two widely adopted RL algorithms. % PPO calculates advantages by estimating state values with the critic model, while GRPO compares the outcomes of multiple rollouts to obtain the advantage values.
% While PPO relies on extrinsic signals from a critic network to estimate action values, GRPO focuses more on intrinsic signals by comparing a batch of outputs from the model policy for each input. 
Given the current policy $\pi_\theta$, query $q$, and the sampled rollout $a$, PPO and GRPO can be formulated as:
\vspace{-1em}

\begin{equation}
\label{eq:ppo_objective}
\small
\resizebox{\textwidth}{!}{$
\mathcal{J}_{\text{PPO}}(\pi_\theta) = \mathbb{E}_{q\sim D, a\sim \pi_\theta(q)} \left[ \frac{1}{|a|}\sum_{j=1}^{|a|} \min\left\{r_{j} A_{j},\ \operatorname{clip}(r_{j}, 1-\epsilon, 1+\epsilon) A_{j}\right\} - \beta\ \mathrm{KL}(\pi_\theta||\pi_\text{ref})\right]$},
\end{equation}

\vspace{-1.5em}

\begin{equation}
\label{eq:ppo_advantage}
\small
\text{where }r_{j} = \frac{\pi_\theta(a_{j} \mid q, a_{1...j-1})}{\pi_{\text{old}}(a_{j} \mid q, a_{1...j-1})},\ A_{j} = Q(q,a_{1...j})-V(q,a_{1...j}).
\end{equation}

\vspace{-0.5em}

\begin{equation}
\label{eq:grpo}
\small
\resizebox{\textwidth}{!}{$
\mathcal{J}_{\text{GRPO}}(\pi_\theta) =
\mathbb{E}_{q\sim D,\, a_{1...N}\sim \pi_\theta(q)}
\left[
\frac{1}{N}\sum_{i=1}^N
\frac{1}{L_\text{max}}\sum_{j=1}^{|a_i|}
\min\left\{
r_{i,j} A_i,\,
\operatorname{clip}(r_{i,j}, 1-\epsilon, 1+\epsilon) A_i
\right\}
-\beta\, \mathrm{KL}(\pi_\theta\|\pi_{\text{ref}})
\right],
$}
\end{equation}

\vspace{-1.5em}

\begin{equation}
\label{eq:advantage}
\small
\text{where } r_{i,j} = \frac{\pi_\theta(a_{i,j} \mid q, a_{i,1...j-1})}{\pi_{\text{old}}(a_{i,j} \mid q, a_{i,1...j-1})},\ A_{i}=R(q,a_i)-\operatorname{mean}[R(q,a_{1...N})].
\end{equation}

In the above formulas, $A$ is the advantage, $D$ is the training dataset, and $r$ refers to the importance sampling ratio. $N$, $\epsilon$, $L_\text{max}$ and $\beta$ are hyperparameters. Specifically, we use Dr. GRPO~\citep{liu2025understanding}, an unbiased variant of GRPO in this paper%\footnote{Dr. GRPO uses $\frac{1}{L_\text{max}}$ instead of $\frac{1}{|a_i|}$ in \Cref{eq:grpo}, and removes the denominator $\operatorname{std}\left[R(q,a_{1...N})\right]$ in advantage calculation (\Cref{eq:advantage}). These changes are adopted as traditional GRPO may lead to unexpected length reduction when efficiency-related rewards are introduced, similar to \citet{arora2025training}.}
. For verifiable tasks, the reward for PPO and GRPO can be obtained by comparing the model answer with the expected ground truth ($gt$):

% with slight differences: Dr. GRPO uses $\frac{1}{L_\text{max}}$ instead of $\frac{1}{|a_i|}$ in \Cref{eq:grpo}, and removes the denominator $\operatorname{std}\left[R(q,a_{1...N})\right]$ in advantage calculation (\Cref{eq:advantage})\footnote{We adopt these two modifications as we found 

% Note that the PPO loss is estimated over a single answer to a given question, while the GRPO loss is estimated over a batch of $N$ answers. 
% Note that the advantage values for GRPO are measured as the normalized reward over the batch of answers, where the verifiable reward for each answer

\vspace{-0.5em}

\begin{equation}
\label{eq:correct_reward}
\small
R_\text{PPO/GRPO}(q,a,gt)=R_\text{VR}(q,a,gt) = 
\begin{dcases} 
1 & \text{if } gt\text{ in } a, \\ 
0 & \text{otherwise}.
\end{dcases}
\end{equation}
Although RL with verifiable reward has set new benchmarks in reasoning tasks, it induces undesired characteristics such as redundant output due to the coarse binary nature of the reward. To address this issue, two widely used methods introduce targeted penalities on output length: O1-pruner~\citep{luo2025o1} and Efficient Reasoner~\citep{arora2025training} (referred to as \textbf{O1} and \textbf{ER}). Their detailed formulations are provided in \Cref{method:baseline}. Despite their effectiveness, both methods inevitably trade off accuracy, which constrains their broader applicability.

\subsection{Self-Aligned Reward}
\label{method:sapo}

% \peixuan{I feel the methodology is a bit short.}
We propose \textbf{Self-Aligned Reward (SAR)} (denoted as $R_\text{SA}$), an unsupervised holistic signal that combines generation quality, conciseness, and query-answer relevance. The self-aligned reward is estimated entirely by the model policy without any external feedback. We then combine our reward with the binary verifiable reward to address the shortcomings of previous methods, aiming to improve both accuracy and generation efficiency. The reward for self-aligned PPO and self-aligned GRPO (\textbf{SA-PPO} and \textbf{SA-GRPO}) can be formulated as follows\footnotemark: 

\begin{equation}
\label{eq:pplqa}
\small
R_\text{SA-PPO/GRPO}(q,a_i,gt)=R_\text{VR}+\alpha R_\text{SA},\ R_\text{SA}=\operatorname{clip}\big(\frac{\operatorname{ppl}(a_i)-\operatorname{ppl}(a_i|q)}{\operatorname{ppl}(a_i)},-1,1\big)
\end{equation}

 \footnotetext{Since $ppl > 0$, we always have $R_\text{SA} < 1$. In practice, $R_\text{SA}$ lies in the $(0,1)$ range in most cases.}

\vspace{-0.5em}

\begin{equation}
\label{eq:ppl}
\small
\text{where }
\operatorname{ppl}(a) = e^{-\frac{1}{|a|}\sum_{j=1}^{|a|} logP(a_j|a_{1...j-1})},\ \operatorname{ppl}(a|q)=e^{-\frac{1}{|a|}\sum_{j=1}^{|a|} logP(a_j|q,a_{1...j-1})}.
\end{equation}

% Since $\operatorname{ppl}(a), \text{ } \operatorname{ppl}(a|q)>0$, the max function ensures $R_\text{SA} \in [-1,1]$\footnote{In practice, we almost always have $R_\text{SA}>0$.}. Thus the overall reward is bounded as follows:

% \begin{equation}
% \label{eq:ppl-bound}
% R_\text{SAPO}(q,a_i,gt) \in 
% \begin{cases}
% [1-\alpha, 1+\alpha] & \text{if } R_\text{VR}(q, a_i) = 1 \\
% [-\alpha, \alpha] & \text{otherwise}.
% \end{cases}
% \end{equation}

The self-aligned reward captures the relative difference between 
$\operatorname{ppl}(a)$ and $\operatorname{ppl}(a|q)$, or the 
\textbf{conditioned perplexity drop}, based on the current model policy. Perplexity reflects the likelihood that the model will produce the given response, with lower values indicating higher confidence. Thus, $R_\text{SA}$ can be interpreted as: 
\textit{``How much less likely does the answer become if the query is not present?''} 

When an answer is tightly tailored to the query, the conditioned perplexity $\operatorname{ppl}(a|q)$ 
will be significantly lower than the standalone perplexity $\operatorname{ppl}(a)$, leading to a 
higher $R_\text{SA}$. In contrast, if parts of the answer are irrelevant
to the query or include noisy, verbose text, the two perplexities will be similar, leading to a smaller $R_\text{SA}$. Therefore, a larger value of $R_\text{SA}$ indicates the answer's stronger dependency and better alignment to the query. 

\Cref{sec:case_study} provides an in-depth analysis of this reward formulation, where we demonstrate that self-aligned reward encourages concise reasoning 
and effective use of query information, thereby improving both accuracy 
and training efficiency, and penalizing noisy verbose answers. In addition, the computation of $R_\text{SA}$ integrates seamlessly into the RL pipeline with negligible computational overhead, as shown in 
\Cref{sec:time,app:implementation}.

\section{Case Analysis: Why Self-aligned Reward Works}
\label{sec:case_study}

This section analyzes the self-aligned reward formulation, providing insights on how SAR enhances the reasoning effectiveness and efficiency.

% In short, SAPO has the following features:

% \textbullet \hspace{1pt} SAPO penalizes completely irrelevant answers, helping the model on track quickly.

% \textbullet \hspace{1pt} SAPO penalizes pulling directly from memorization, promoting reasoning.

% \textbullet \hspace{1pt} SAPO promotes shorter and concise answers over long and redundant ones.

% \textbullet \hspace{1pt} SAPO promotes using information from the question, making the answer more grounded on the question.

\subsection{SAR Provides Accurate and Fine-grained Signals}

% Correctness reward (\Cref{eq:correct_reward}) only provides binary signals, failing to give fine-grained judgements among correct or incorrect answers.

An ideal reward design should be able to rate different types of answers based on their qualities. To assess different reward functions, we analyze 6 different types of answers to 200 math questions sampled from five common math benchmarks (\cref{subsec:exp_settings}) in \Cref{table:category}. Types \textbf{(1)} to \textbf{(4)} are obtained by sampling Qwen3-1.7B rollouts at temperature $=1$, and using GPT-4o to annotate the responses (see \Cref{app:prompt} for details). Types \textbf{(5)} and \textbf{(6)} are artificially synthesized to simulate memorization, where the LLM directly extracts answers from its knowledge without reasoning.

% \textbullet \hspace{1pt} \textbf{(1)}: Correct and concise answers, which is most preferable;

% \textbullet \hspace{1pt} \textbf{(2)}: Correct but wordy answers that contains unnecessary content;

% \textbullet \hspace{1pt} \textbf{(3)}: Partly correct answers, which makes some mistake and fails to reach the final ground truth;

% \textbullet \hspace{1pt} \textbf{(4)}: Completely wrong or irrelevant answers, indicating the model doesn't understand the problem.

% We also artificially synthesize two other types of answers\footnote{Lengths for type \textbf{(5)} and \textbf{(6)} are defined as 50.}: 

% \textbullet \hspace{1pt} \textbf{(5)}: Correct answer with no thinking process; 

% \textbullet \hspace{1pt} \textbf{(6)}: Incorrect answer with no thinking process. Type \textbf{(5)} and \textbf{(6)} simulates memorization where the LLM directly extracts answers from its knowledge without reasoning.

% \adit{For the table change all rewards to advantage value including correctness reward}
\begin{table}[htbp]
\caption{We calculate advantage values over the 6 types of answers to the same question ($\alpha=1$), and report the average over 200 questions. \textbf{The advantage values of SA-GRPO accurately rate answers of different qualities.}}
\label{table:category}
\small
\centering
\renewcommand{\arraystretch}{1.1}
\setlength{\tabcolsep}{4pt} % Reduces space to 3pt
\resizebox{0.85\textwidth}{!}{%
\begin{tabular}{lccccc}
\toprule
Answer Type                      & Length & $A_\text{GRPO}$ & $A_\text{GRPO-o1}$ & $A_\text{GRPO-ER}$ & $A_\text{SA-GRPO}$ (Ours) \\\hline
\textbf{(1)}: Correct and concise              &    143.5    &      0.5       &     1.04      & 0.81   &    1.15  \\
\textbf{(2)}: Correct but redundant            &   236.0     &      0.5       &      0.35     &  -0.11  &    1.00  \\
\textbf{(3)}: Partly correct with wrong answer &    457.1    &      -0.5       &     -1.28      &  -0.65  &   -0.01   \\
\textbf{(4)}: Completely wrong or irrelevant   &     405.8   &      -0.5       &      -1.19     &  -0.65  &   -1.44   \\
\textbf{(5)}: Correct but no thought               &   5.0     &      0.5      &     1.04      &  1.25  &  -0.04    \\
\textbf{(6)}: Incorrect and no thought            &   5.0     &      -0.5       &     0.04      & -0.65   &  -0.68   \\\bottomrule
\end{tabular}}
\end{table}

% \begin{table}[t]
% \caption{Example}
% \label{table:category_case}
% \small
% \centering
% \renewcommand{\arraystretch}{1.1}
% \setlength{\tabcolsep}{4pt} % Reduces space to 3pt
% \begin{tabular}{lccccc}
% \toprule
% Answer Type                      & Length & $R_\text{cor}$ & $R_\text{o1}$ & $R_\text{ER}$ & $R_\text{SA}$ (Ours) \\\hline
% \textbf{(1)}: Correct and concise              &        &      1.0       &           &    &      \\
% \textbf{(2)}: Correct but redundant            &        &      1.0       &           &    &      \\
% \textbf{(3)}: Partly correct with wrong answer &        &      0.0       &           &    &      \\
% \textbf{(4)}: Completely wrong or irrelevant   &        &      0.0       &           &    &      \\
% \textbf{(5)}: Correct but no thought               &        &      1.0      &           &    &      \\
% \textbf{(6)}: Incorrect and no thought            &        &      0.0       &           &    &     \\\bottomrule
% \end{tabular}
% \end{table}

We present the advantage values (\Cref{eq:advantage}) for each type of response using the following rewards: verifiable GRPO rewards ($A_\text{GRPO}$), length-based rewards from the O1 and ER methods ($A_\text{GRPO-o1}$, $A_\text{GRPO-ER}$) and our self-aligned reward $A_\text{SA-GRPO}$. From \Cref{table:category}, we can observe that SAR provides rich and accurate signals to different answers, exhibiting the following features:

\textbullet \hspace{1pt} SAR favors concise and correct answers; it gives a lower reward to long and redundant answers, promoting efficiency. \Cref{sec:case2} explains this phenomenon in detail.

\textbullet \hspace{1pt} SAR provides partial credit to partly correct answers and penalizes completely irrelevant ones, helping the model learn basic reasoning patterns in the initial stage of training.

\textbullet \hspace{1pt} SAR penalizes the synthesized ``no thought'' answers, even if they're correct and short. This indicates that the reasoning process plays a critical role in SAR, and memorization is discouraged.

On the other hand, the verifiable reward (GRPO) fails to discern answers that are both correct and incorrect, and O1 or ER focus solely on response length, making signals biased and not suitable for maximizing accuracy. For instance, ER isn't applicable to wrong answers, and O1 favors irrelevant answers over partly correct ones, simply because they're shorter. We also provide a qualitative example in \Cref{table:category_case} to illustrate the different answer types.

\subsection{SAR Promotes Exploiting Query Information Efficiently}
\label{sec:case2}

In this experiment, we take a deeper look at SAR at the token level, aiming to reveal which tokens contribute more to the overall score. Specifically, we decompose the self-aligned reward  (\Cref{eq:pplqa,eq:ppl}) to calculate the token-level score\footnote{We omit the clipping for simplicity.}:
\begin{equation}
\small
\label{eq:pplqa_decompose}
R_\text{SA}=1-\frac{\operatorname{ppl}(a|q)}{\operatorname{ppl}(a)}=1-e^{-\frac{1}{|a|}\sum_{j=1}^{|a|} log\frac{P(a_j|q,a_{1...j-1})}{P(a_j|a_{1...j-1})}}.
\end{equation}

We then define $v(a_j)=log\frac{P(a_j|q,a_{1...j-1})}{P(a_j|a_{1...j-1})}$ to measure the importance of each token $a_j$ on the whole metric $R_\text{SA}$. A token with a higher $v(a_j)$ is considered valuable, while a token with a low or even negative $v(a_j)$ indicates it's less informative and independent of the query.

\begin{figure}[htbp]
    \centering
    \includegraphics[width=0.85\linewidth]{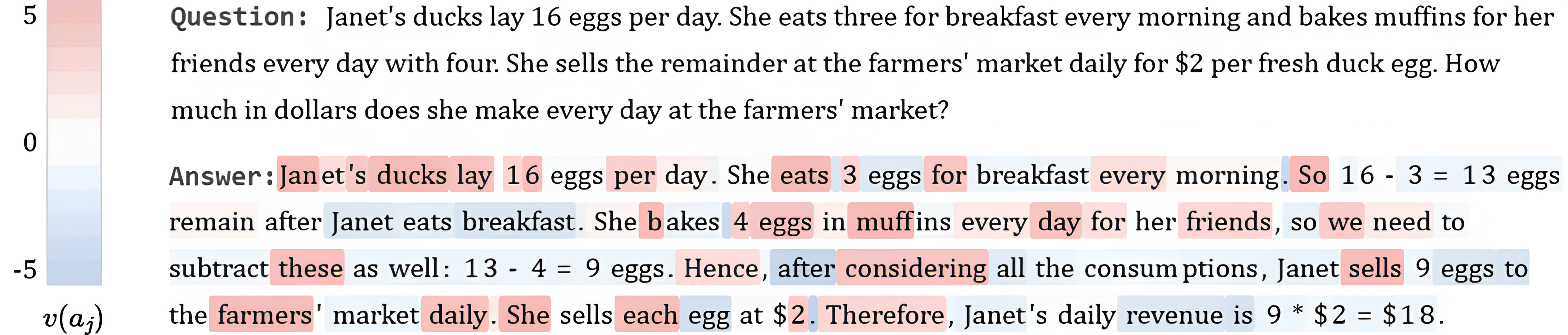}
    \caption{An illustration of token-level importance scores (i.e. $v(a_j)$). Red color means the token is considered informative for our self-aligned reward, and blue indicates a negative $v(a_j)$. \textbf{Tokens extracting new information from the query get high scores.}}
    \label{fig:token_level_case}
\end{figure}

From \Cref{fig:token_level_case}, we observe that tokens drawing on information from the question for the first time, like ``Janet'', ``duck'', ``16'', tend to receive high scores. This is because such information is present in the query but absent from previous answer tokens, making $P(a_j|q,a_{1...j-1})$ relatively high while $P(a_j|a_{1...j-1})$ remains low. In contrast, repeating information already generated, like mentioning ``Janet'' for the second time, results in low scores as both probabilities become similarly high.

Generally, tokens in the earlier part of an answer typically achieve higher $v(a_j)$ values, since extracting new content from the query is easier at the beginning. Later tokens, by comparison, struggle to introduce novel information once much of the query has already been incorporated. This explains why SAR favors short, concise answers and promotes more efficient generation.

% We can observe the following phenomenon from \Cref{fig:token_level_case}:

% \textbullet \hspace{1pt} Tokens that utilize information in the question for the first time will get a high score, like the numbers ``16'', ``3'', ``4''. Since such information can only be found in the query but not in previous answer tokens, $P(a_j|q,a_{1...j-1})$ is relatively high and $P(a_j|a_{1...j-1})$ is very low. This feature helps the model focus on the exact information in the query, rather than hallucinating or generating information aimlessly.

% \textbullet \hspace{1pt} Repeating information that has appeared before leads to a low score. Since such information can be found in previous answer tokens, both $P(a_j|q,a_{1...j-1})$ and $P(a_j|a_{1...j-1})$ are equally high. This feature penalizes redundancy and repetition.

% \textbullet \hspace{1pt} Tokens in the early part generally have higher $v(a_j)$ values than later ones, as extracting new query information is easier in the early stage. In the latter part, most information has already appeared in $a_{1...j-1}$, making it harder to exploit new content from the query. This explains why SAR favors short and concise answers, making the model more efficient.
\section{Experiments}
\label{sec:exp}

\subsection{Experimental Settings}
\label{subsec:exp_settings}

\textbf{Datasets.} We utilize five math reasoning benchmarks: \underline{GSM8k}~\citep{cobbe2021training}, \underline{MATH}~\citep{hendrycks2021measuring}, \underline{NuminaMath 1.5}~\citep{li2024numinamath}, \underline{GSM-symbolic}~\citep{mirzadeh2024gsm}, and \underline{AIME 1983--2024}~\citep{veeraboina_aime_2024}, covering a difficulty range from primary school to competition level. For training, we combine the training splits of the first three datasets, while the remaining two are not present during training to evaluate generalization.

\textbf{Models.} We utilize four base models: \underline{Qwen3-1.7B} (Base), \underline{Qwen3-4B} (Base)~\citep{yang2025qwen3}, \underline{Phi-3.5-mini} (Instruct, 3.8B)~\citep{abdin2024phi}, and \underline{Gemma3-1B} (Instruct)~\citep{team2025gemma}, covering different sizes and model families. % We preferred the base versions of each model for unbiased alignment, where available.

\textbf{Settings.} We train reasoning models with the following methods:
% We compare SAPO against the following baselines:

% \textbullet \hspace{1pt} \texttt{Base}: The untrained model.

% \textbullet \hspace{1pt} \texttt{SFT}: We sample 8 rollouts per question from the base model and select the shortest correct rollout for each question. These question-answer pairs are used to finetune the model.

\textbullet \hspace{1pt} PPO, GRPO: two traditional RL algorithms, using only the verifiable reward.

% \textbullet \hspace{1pt} \texttt{GRPO+Prompt}: Instruct the \texttt{GRPO} model to shorten the reasoning process in the prompt.

\textbullet \hspace{1pt} GRPO-O1, GRPO-ER: algorithms with length penalties, aimed to enhance efficiency. See \Cref{method:baseline} for details.

\textbullet \hspace{1pt} SA-PPO, SA-GRPO: algorithms using \textbf{self-aligned reward (SAR)}. See \Cref{method:sapo} for details.

For training details and hyperparameters, refer to \Cref{app:training}.

\textbf{Metrics.} In this work, we focus on accuracy and efficiency (measured by average answer length) of LLM reasoning. We report these two metrics as well as an Accuracy-Efficiency trade-off Score (AES). For a trained policy $\pi_\theta$ and its base model $\pi_\text{ref}$, we define $\Delta \operatorname{len}=\frac{\operatorname{len}(\pi_\text{ref})-\operatorname{len}(\pi_\theta)}{\operatorname{len}(\pi_\text{ref})}$ and $\Delta \operatorname{acc}=\frac{\operatorname{acc}(\pi_\theta)-\operatorname{acc}(\pi_\text{ref})}{\operatorname{acc}(\pi_\text{ref})}$. Then, $\text{AES}(\pi_\theta)=\Delta \operatorname{len}+\gamma \Delta \operatorname{acc}$ measures the trade-off.\footnote{The choice of $\gamma$ value reflects which aspect does the user prioritize. Since accuracy is the most important factor in most use cases, we set $\gamma=5$ in practice.}

\subsection{Main Results}
\label{sec:exp_result}

\definecolor{mypurple}{RGB}{138, 71, 205}

\begin{table}[t]
\caption{Evaluation on math benchmarks. \textbf{Self-aligned reward significantly reduces token usage while achieving the best reasoning accuracy}. GRPO-O1 and GRPO-ER results in the table use $\alpha=0.05$, and SA-GRPO uses $\alpha=0.2$. Best results among each model are bolded.}
\vspace{0.4em}
\label{table:result}
\small
\centering
\renewcommand{\arraystretch}{1.05}
\setlength{\tabcolsep}{4pt} % Reduces space to 3pt
\resizebox{0.96\textwidth}{!}{%
\begin{tabular}{lcccccccccc|ccc}
\toprule
\multirow{2}{*}{\textbf{Setting}} & \multicolumn{2}{c}{GSM8k} & \multicolumn{2}{c}{GSM-sym} & \multicolumn{2}{c}{MATH} & \multicolumn{2}{c}{NuminaMath} & \multicolumn{2}{c}{AIME} & \multicolumn{3}{c}{\textbf{Average}} \\
           & acc        & len        & acc         & len        & acc          & len         & acc          & len          & acc         & \multicolumn{1}{c}{len}        & \multicolumn{1}{c}{acc}     & len     & AES     \\ \hline\hline
\rowcolor{gray!15}
\textbf{Qwen3-1.7B} & 69.22 & 281 & 46.76 & 365 & 56.89 & 700 & 24.58 & 1242 & 9.97 & 1539 & 41.48 & 825.4 & 0.000 \\
% ~~~+SFT       & 73.01 & 263 & 47.90 & 351 & 58.42 & 668 & 26.26 & 1205 & 8.57 & 1471 & 42.83 & 791.6 & 0.204 \\

~~~+PPO       & 82.71 & 332 & 62.88 & 460 & 63.04 & 784 & 27.02 & 1358 & 9.11 & 1718 & 48.95 & 930.4 & 0.773 \\

\rowcolor{mypurple!7}
~~~\textbf{+SA-PPO}    & \textbf{85.51} & 281 & \textbf{68.02} & 369 & 66.27 & 587 & 32.05 & 866 & 10.61 & 1015 & 52.49 & 623.6 & 1.572 \\
~~~+GRPO      & 84.53 & 335 & 67.66 & 413 & 67.31 & 697 & 33.44 & 1091 & 13.93 & 1278 & 53.37 & 762.8 & 1.509 \\

~~~+GRPO-O1   & 84.15 & 285 & 65.78 & 363 & 66.51 & 528 & 33.15 & 775 & 13.61 & 910 & 52.64 & 572.2 & 1.652 \\
~~~+GRPO-ER   & 82.71 & 244 & 64.74 & 320 & 66.54 & 557 & 34.35 & 980 & 12.22 & 1232 & 52.11 & 666.6 & 1.474 \\
\rowcolor{mypurple!7}
~~~\textbf{+SA-GRPO}      & \textbf{85.51} & 267 & 67.66 & 346 & \textbf{67.96} & 564 & \textbf{36.03} & 841 & \textbf{13.50} & 992 & \textbf{54.13} & 602.0 & \textbf{1.795} \\ \noalign{\vskip 2pt}
\rowcolor{gray!15}
\textbf{Qwen3-4B}   & 76.19 & 315 & 66.22 & 435 & 63.79 & 652 & 38.80 & 1142 & 20.15 & 1281 & 53.03 & 765.0 & 0.000 \\
% ~~~+SFT       & 79.38 & 261 & 66.42 & 362 & 68.39 & 606 & 39.53 & 1017 & 19.19 & 1184 & 54.58 & 686.0 & 0.249 \\

~~~+PPO       & 91.36 & 373 & 85.42 & 485 & 82.15 & 1132 & 50.80 & 1995 & 26.90 & 2680 & 67.33 & 1333.0 & 0.606 \\
\rowcolor{mypurple!7}
~~~\textbf{+SA-PPO}    & 92.12 & 266 & 83.52 & 345 & 78.67 & 652 & 49.09 & 1133 & 28.19 & 1397 & 66.32 & 758.6 & 1.260 \\
~~~+GRPO      & 92.62 & 320 & 86.30 & 414 & 82.51 & 821 & 53.79 & 1485 & 30.11 & 2113 & 69.07 & 1030.6 & 1.165 \\

~~~+GRPO-O1   & 91.89 & 262 & 86.76 & 350 & 80.41 & 549 & 51.06 & 866 & 28.62 & 1016 & 67.75 & 608.6 & 1.592 \\
~~~+GRPO-ER   & 92.80 & 219 & 85.24 & 297 & 79.48 & 503 & 50.47 & 854 & 28.94 & 1022 & 67.38 & 579.0 & \textbf{1.596} \\
\rowcolor{mypurple!7}
~~~\textbf{+SA-GRPO}      & \textbf{93.40} & 239 & \textbf{87.64} & 323 & \textbf{82.63} & 762 & \textbf{57.70} & 1358 & \textbf{35.69} & 1788 & \textbf{71.41} & 894.0 & 1.564 \\ \noalign{\vskip 2pt}

\rowcolor{gray!15}
\textbf{Phi-3.5-mini}   & 71.19 & 287 & 56.86 & 398 & 33.83 & 1132 & 11.56 & 1492 & 2.35 & 943 & 35.15 & 850.4 & 0.000 \\
% ~~~+SFT       & 73.69 & 236 & 35.10 & 797 & 23.28 & 1022 & 7.70 & 1064 & 2.04 & 883 & 28.36 & 800.4 & -0.907 \\

~~~+PPO       & 82.49 & 209 & 64.88 & 322 & 38.56 & 831 & 11.45 & 1707 & \textbf{3.85} & 926 & 40.24 & 799.0 & 0.784 \\
\rowcolor{mypurple!7}
~~~\textbf{+SA-PPO}    & 87.04 & 238 & 73.52 & 344 & 46.34 & 300 & \textbf{17.43} & 586 & 2.79 & 501 &  45.42& 393.8 &  1.997 \\
~~~+GRPO      & 87.56 & 235 & \textbf{73.76} & 337 & 48.69 & 415 & 16.37 & 582 & 3.21 & 672 & 45.91 & 448.2 & 2.003 \\

~~~+GRPO-O1   & 79.83 & 203 & 62.24 & 279 & 44.41 & 390 & 14.08 & 509 & 2.57 & 583 & 40.62 & 392.8 & 1.316\\
~~~+GRPO-ER   & 82.03 & 154 & 63.54 & 240 & 43.61 & 363 & 12.55 & 511 & 2.47 & 606 & 40.84 & 374.8 & 1.368\\
\rowcolor{mypurple!7}
~~~\textbf{+SA-GRPO}      & \textbf{87.95} & 207 & 72.72 & 289 & \textbf{50.99} & 356 & 16.08 & 455 & 3.22 & 535 & \textbf{46.19} & 368.4 & \textbf{2.137} \\ \noalign{\vskip 2pt}

\rowcolor{gray!15}
\textbf{Gemma3-1B}   & 42.15 & 325 & 19.58 & 427 & 36.40 & 1042 & 12.40 & 1414 & 2.25 & 2279 & 22.56 & 1097 & 0.000 \\
% ~~~+SFT       & 35.41 & 293 & 11.86 & 407 & 27.04 & 1060 & 10.80 & 1408 & 1.39 & 2182 & 17.30 & 1070.0 & -1.141 \\

~~~+PPO       & 56.86 & 942 & 30.28 & 1187 & 44.17 & 1408 & 15.57 & 1904 & 1.60 & 2422 & 29.69 & 1572.6 & 1.146\\
\rowcolor{mypurple!7}
~~~\textbf{+SA-PPO}    & 55.80 & 683 & 31.00 & 936 & 42.94 & 1025 & 14.22 & 1523 & 1.50 & 1683 & 29.10 & 1170.0 & 1.383 \\
~~~+GRPO      & 59.97 & 1208 & \textbf{34.70} & 1552 & 45.40 & 1693 & 16.59 & 2264 & 2.25 & 2613 & 31.78 & 1866.0 & 1.343 \\

~~~+GRPO-O1   & 60.80 & 873 & 32.06 & 1157 & 44.76 & 1298 & 16.16 & 1871 & \textbf{2.35} & 2449 & 31.23 & 1529.6 & 1.528 \\
~~~+GRPO-ER   & 59.44 & 748 & 32.36 & 1063 & 46.23 & 1131 & 16.41 & 1545 & 2.04 & 1696 & 31.29 & 1236.6 & 1.808 \\
\rowcolor{mypurple!7}
~~~\textbf{+SA-GRPO}      & \textbf{61.26} & 552 & 34.52 & 772 & \textbf{46.60} & 952 & \textbf{16.70} & 1302 & 2.14 & 1509 & \textbf{32.24} & 1017.4 & \textbf{2.218} \\
\bottomrule
\end{tabular} }
\end{table}

% \footnotetext{We exclude the AIME dataset for Phi-3.5-mini and Gemma3-1B when \textbf{calculating the average}, as the accuracy under all settings are $\leq 3\%$.}

From \Cref{table:result}, we observe that \textbf{baseline approaches struggle to balance accuracy and efficiency.} PPO and GRPO significantly improve accuracy at the cost of longer and possibly redundant answers. While GRPO-O1 and GRPO-ER effectively reduce computation overhead, they compromise reasoning capability. In contrast, \textbf{SAR delivers substantial improvements in both accuracy and efficiency}. Across four base models, SA-GRPO consistently achieves the highest reasoning accuracy while maintaining highly efficient reasoning, with at least 4\% improvement in accuracy and 30\% reduction in length compared to GRPO. Notably, SA-GRPO produces answers of comparable or even shorter length than GRPO-O1 and GRPO-ER, which are explicitly designed for efficient reasoning. A similar phenomenon is observed for SA-PPO, indicating that self-aligned reward applies to diverse RL algorithms. We show two examples comparing GRPO and SA-GRPO in \Cref{app:case}.

These findings highlight the superiority of the self-aligned reward mechanism. With a more fine-grained and intelligent signal, SAR makes the model preserve the necessary reasoning which are closely related to query information, and reduces unnecessary content that introduces token burdens, leading to more effective and efficient reasoning.

\subsection{Self-aligned Reward Achieves a Pareto-Optimal in Dynamic Balance}
\label{exp_tradeoff}

% \adit{Update other plots to the format of the Qwen 1.7B plot. Also need to change the shape of Gemma 1B plot}
\begin{figure}[htbp]
    \centering
    \subcaptionbox{Qwen3-1.7B.\label{fig:a}}[0.48\textwidth]{
        \includegraphics[width=\linewidth]{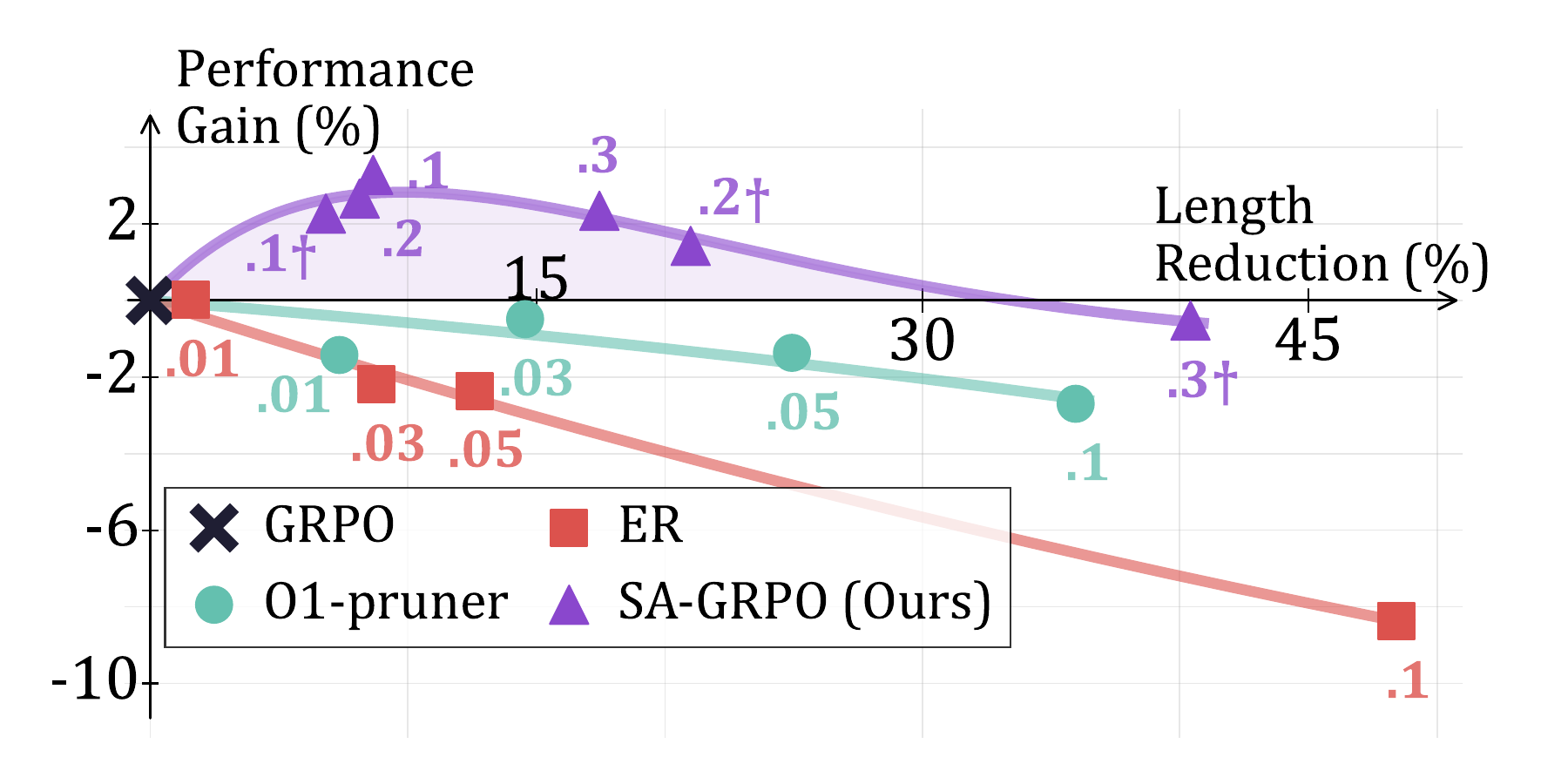}
    }
    \hfill
    \subcaptionbox{Qwen3-4B.\label{fig:b}}[0.48\textwidth]{
        \includegraphics[width=\linewidth]{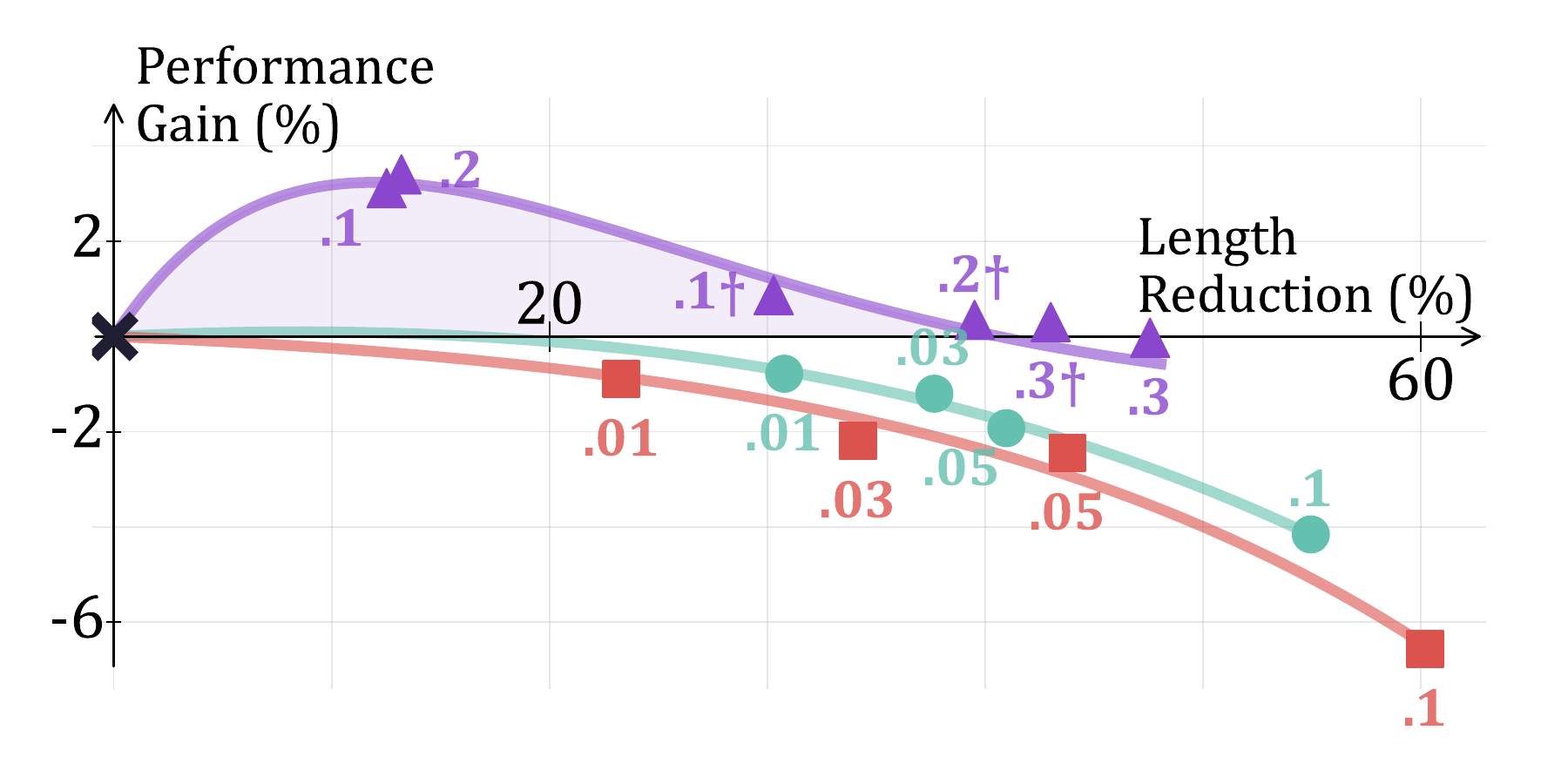}
    }
    
    \subcaptionbox{Phi-3.5-mini.\label{fig:c}}[0.48\textwidth]{
        \includegraphics[width=\linewidth]{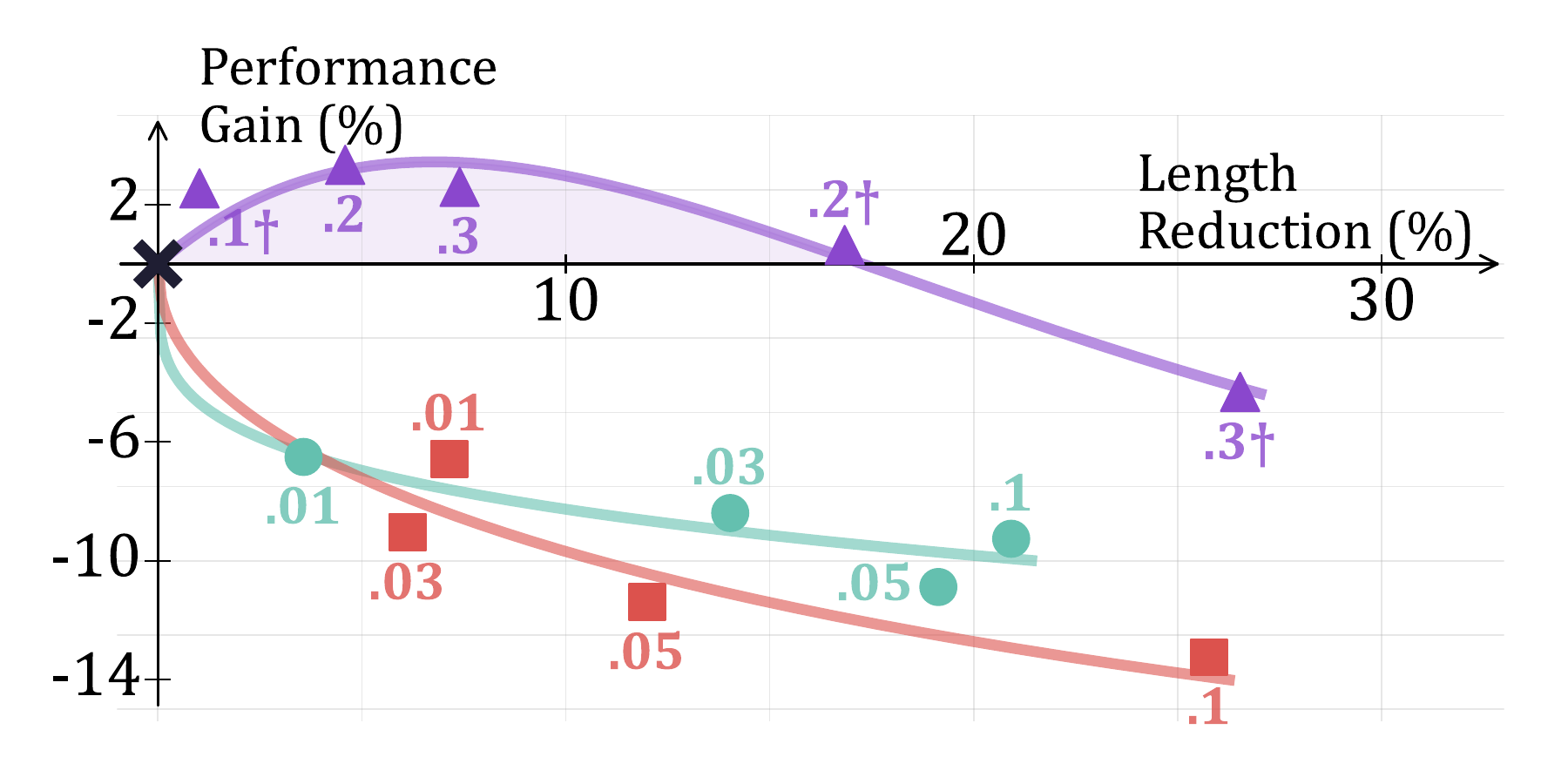}
    }
    \hfill
    \subcaptionbox{Gemma3-1B.\label{fig:d}}[0.48\textwidth]{
        \includegraphics[width=\linewidth]{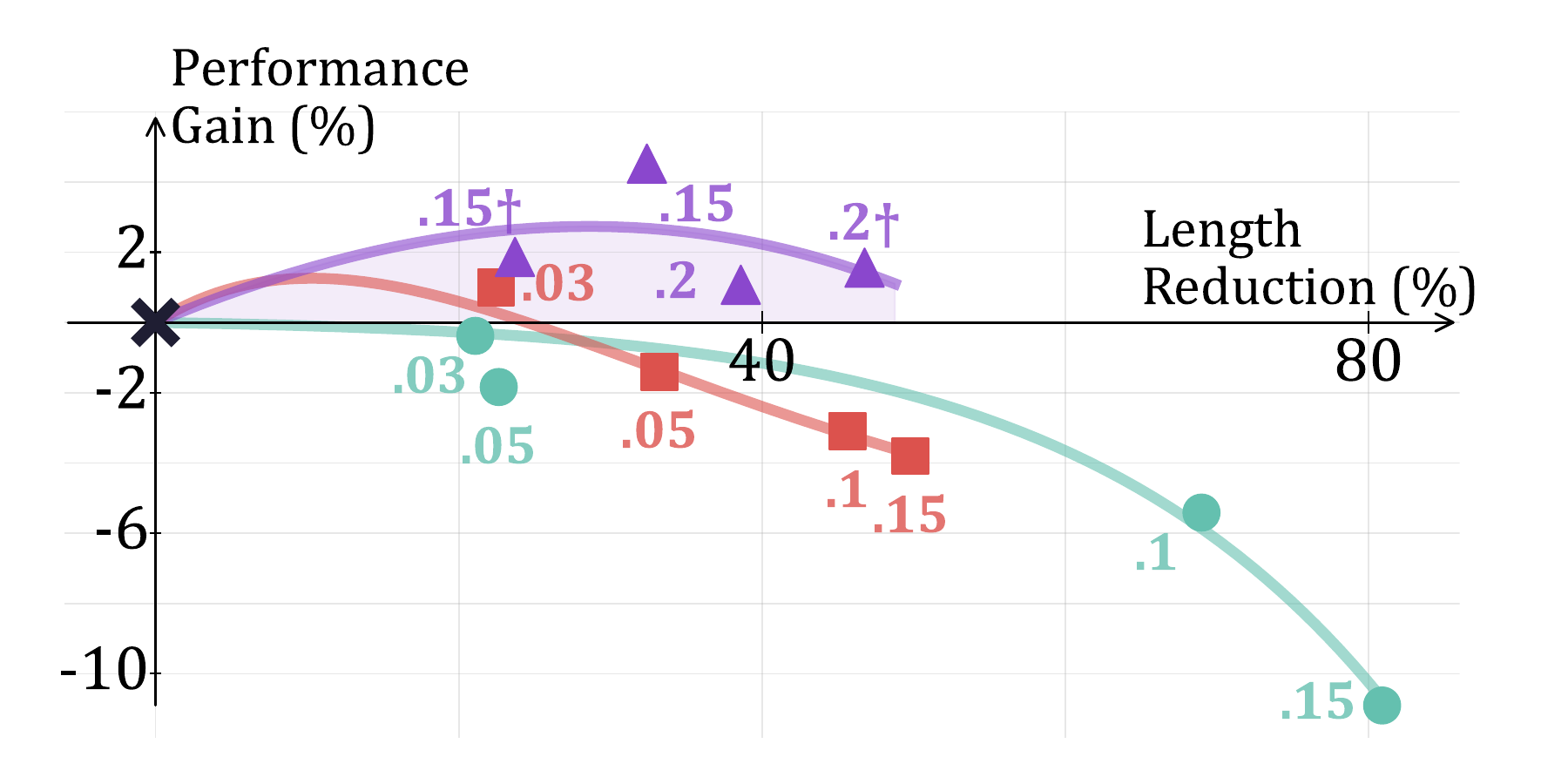}
    }
    \caption{Accuracy-efficiency balance of different algorithms. \textbf{SA-GRPO reaches a Pareto-optimal curve and achieves notable gains on both axes}. Numbers around each point indicate the $\alpha$ values, and † indicates using 20\% of the total training data with the same total training steps.}
    \label{fig:contours}
\end{figure}

The hyperparameter $\alpha$ in \Cref{eq:pplqa} controls the weights of verifiable reward and self-aligned reward, offering flexibility to focus more on efficiency or accuracy. Therefore, we train SA-GRPO with different $\alpha$ values and compare it with GRPO-O1 and GRPO-ER on 2-dimensional plots.

From \Cref{fig:contours}, we can observe that the curves for SA-GRPO are consistently on the top-right side over curves for GRPO-O1 and GRPO-ER, indicating SA-GRPO balances accuracy and efficiency better than length-based methods. Moreover, length-based methods are always under the x-axis in most cases, showing that these methods sacrifice accuracy for reduced tokens, while SA-GRPO is the only method consistently above the x-axis (illustrated as the light purple area in the figure), showing better accuracy and better efficiency compared to vanilla GRPO. The Pareto-optimal balance between accuracy and efficiency of SA-GRPO further demonstrates the effectiveness of SAR.

% Interestingly, different base models lead to different curve shapes and scales. For instance, length-based methods have concave curves for \Cref{fig:b,fig:c}, but these curves are convex for \Cref{fig:a,fig:d}. In addition, the shape of SA-GRPO curves is also model-dependent. This indicates the base model has a large impact on RL behaviors~\citep{zeng2025simplerl}.

\subsection{Self-aligned reward Generalizes to Logical Reasoning}
\label{exp:logic}

In this section, we examine the generalization ability of self-aligned reward (SAR) to a different domain — logical reasoning - via two benchmark datasets: LogicBench~\citep{parmar2024logicbench} and ProntoQA~\citep{saparov2022language}. In \Cref{table:logic}, we  find that SA-GRPO improves accuracy compared to GRPO on 5 out of 8 columns, and outperforms length-based methods (GRPO-O1 and ER) on all cases. Similarly, SA-PPO outperforms the PPO baseline on 6 columns. SAR maintains an efficiency gain in the logical domain, not only reducing answer length compared to GRPO and vanilla length-based methods in most cases. These results show that SAR generalizes well to the out-of-domain logical reasoning benchmarks, validating our overall reward design.

\begin{table}[t]
\caption{\textbf{SA-PPO and SA-GRPO maintains robustness and efficiency in out-of-domain tasks.}}
\vspace{0.4em}
\label{table:logic}
\footnotesize
\centering
\renewcommand{\arraystretch}{1.2}
\setlength{\tabcolsep}{3pt} % Reduces space to 3pt

\begin{subtable}[t]{0.49\textwidth}
\centering
\caption*{(a) LogicBench.}
\resizebox{!}{6.5em}{
\begin{tabular}{lcccccccc}
\toprule
\textbf{Model} &
\multicolumn{2}{c}{\shortstack{Qwen3\\-1.7B}} &
\multicolumn{2}{c}{\shortstack{Qwen3\\-4B}} &
\multicolumn{2}{c}{\shortstack{Phi-3.5\\-mini}} &
\multicolumn{2}{c}{\shortstack{Gemma3\\-1B}} \\
              & acc & len & acc & len & acc & len & acc & len \\ \midrule
              \rowcolor{gray!15}
Base & 56.0 & 182 & 75.4 & 336 & 66.0 & 309 & 50.6 & 303 \\
% SFT          &    56.0  &  187   &   77.0   &  319   &  63.2    &   271  &   42.6   & 223    \\

PPO          &   60.2   &  340   &   80.0   &   455  &    60.8  &  229   &    48.8  &  264   \\
\rowcolor{mypurple!7}
\textbf{SA-PPO}          &  64.6    &  368   &  77.6  &  342  &  64.2  &  205  & \textbf{55.6} & 241     \\

GRPO         &   64.2   &  372   &  78.4    &   416  &   \textbf{67.6}  &  243   &  53.0    &  277   \\

GRPO-O1           &  60.6    &  264   &   76.0   &   350  &  51.0    &  212   &  52.6    & 223    \\
GRPO-ER           &  61.4    &   206  &   77.6   &   263  &   49.8   &  143   &  50.0    &   250  \\
\rowcolor{mypurple!7}
\textbf{SA-GRPO}         &   \textbf{65.0}   &  278   &  \textbf{80.4}    &  208   &  57.4    &  173   &  55.2    &  222   \\
\bottomrule
\end{tabular} }
\end{subtable}
\hfill
\begin{subtable}[t]{0.49\textwidth}
\centering
\caption*{(b) ProntoQA.}
\resizebox{!}{6.5em}{
\begin{tabular}{lcccccccc}
\toprule
\textbf{Model} &
\multicolumn{2}{c}{\shortstack{Qwen3\\-1.7B}} &
\multicolumn{2}{c}{\shortstack{Qwen3\\-4B}} &
\multicolumn{2}{c}{\shortstack{Phi-3.5\\-mini}} &
\multicolumn{2}{c}{\shortstack{Gemma3\\-1B}} \\
              & acc & len & acc & len & acc & len & acc & len \\ \midrule
              \rowcolor{gray!15}
Base & 69.0 & 616 & 89.6 & 988 & 94.8 & 339 & 56.2 & 712 \\
% SFT          &   68.0   &  610   &   95.6   &   401  &  97.2    &  331   &  55.8   &  420    \\

PPO          &   71.8   &  920   &    97.8  &   561  &    96.4  &   301  &   56.4   &  904   \\
\rowcolor{mypurple!7}
\textbf{SA-PPO}          &   65.6   &  735   &  99.0  &  405  &  \textbf{98.2}  &  304  & 57.6 &  690   \\
GRPO         &    74.0  &   756  &   \textbf{100.0}   &  546   &   \textbf{98.2}   &  254   &    57.4  &  963   \\

GRPO-O1           &  73.0    &  505   &  98.8    &  469   &   87.2   & 227    & 59.0     &  938   \\
GRPO-ER           &    73.6  &   424  &  99.0    &   309  &  86.6    &   205  &   58.6   & 919    \\
\rowcolor{mypurple!7}
\textbf{SA-GRPO}        &  \textbf{77.8}    &  482   &  99.8    &  301   &  90.8    &  216   &   \textbf{60.6}   &  733   \\
\bottomrule
\end{tabular} }
\end{subtable}

\end{table}

\section{Analysis}
\label{sec:analysis}

\subsection{Ablation Study}
\label{sec:ablation}

In this section, we investigate two critical components of our algorithm that make internal signals effective during training. %, overcoming the challenges discussed in \citet{zhang2025no}. 
Firstly, the self-aligned reward (\Cref{eq:pplqa}) measures \textbf{conditioned perplexity drop}, which is the relative difference between $ppl(a|q)$ and $ppl(a)$, as a proxy for query-answer relevancy. However, existing methods focus mainly on $ppl(a|q)$ alone, which measures entropy or uncertainty\citep{zhao2025learning,agarwal2025unreasonable}.% Through computing the \textbf{conditioned perplexity drop}, the self-aligned reward the coherence between question and answer instead of self-confidence. % Therefore, SA-GRPO avoids overconfidence and keeps the entropy drop at a reasonable rate, promoting diverse exploration.
Secondly, our approach combines \textbf{verifiable reward} with self-aligned reward. % As verifiable rewards are easily obtainable in the math domain, we utilize this information instead of completely removing the term. Combining external and internal signals ensures that answer correctness is always a key component of the reward, maintaining the reasoning quality and training stability.
We ablate these components and derive the following rewards to compare against SA-GRPO ($R=R_\text{VR}+\alpha R_\text{SA}$):

\textbullet \hspace{1pt} $R=R_\text{SA}$: this reward removes the \textbf{ verifiable reward}, using only the self-aligned reward.

\textbullet \hspace{1pt} $R=R_\text{VR}+\alpha R_\text{EM}=R_\text{VR}-\alpha log\operatorname{ppl}(a|q)$: this reward uses ``entropy minimization'', a metric measuring self-confidence, as the internal reward, instead of \textbf{conditioned perplexity drop};

\textbullet \hspace{1pt} $R=R_\text{EM}=-log\operatorname{ppl}(a|q)$: this reward is the entropy minimization reward~\citep{agarwal2025unreasonable}, ablating both components.

\begin{table}[hbtp]
\caption{GRPO training results on Qwen3-4B with different internal reward signals ($\alpha=0.2$). \textbf{The SA-GRPO formulation obtains optimal performance among baselines.}}
\vspace{0.4em}
\label{table:ablation}
\small
\centering
\renewcommand{\arraystretch}{1.05}
\setlength{\tabcolsep}{4pt} % Reduces space to 3pt

\resizebox{0.9\textwidth}{!}{%
\begin{tabular}{lcccccccccc|cc}
\toprule
\multirow{2}{*}{Setting} & \multicolumn{2}{c}{GSM8k} & \multicolumn{2}{c}{GSM-sym} & \multicolumn{2}{c}{MATH} & \multicolumn{2}{c}{NuminaMath} & \multicolumn{2}{c}{AIME} & \multicolumn{2}{c}{Average} \\
           & acc        & len        & acc         & len        & acc          & len         & acc          & len          & acc         & \multicolumn{1}{c}{len}        & \multicolumn{1}{c}{acc}     & len     \\ 
\midrule \midrule
Base   & 76.19 & 315 & 66.22 & 435 & 63.79 & 652 & 38.80 & 1142 & 20.15 & 1281 & 53.03 & 765.0 \\
$R_\text{VR}$      & 92.62 & 320 & 86.30 & 414 & 82.51 & 821 & 53.79 & 1485 & 30.11 & 2113 & 69.07 & 1030.6  \\ \hline
$R_\text{EM}$    & 79.15 & 319 & 67.94 & 414 & 68.89 & 970 & 39.86 & 1885 & 21.44 & 2556 & 55.46 & 1228.8 \\
$R_\text{SA}$    & 39.65 & 87 & 19.54 & 102 & 24.72 & 74 & 16.48 & 80 & 4.39 & 79 & 20.96 & 84.4 \\
$R_\text{VR}+\alpha R_\text{EM}$  & 92.25 & 351 & 87.36 & 414 & 81.83 & 779 & 54.78 & 1371 & 33.01 & 1767 & 69.85 & 936.4 \\
\rowcolor{mypurple!15}
\textbf{$R_\text{VR}+\alpha R_\text{SA}$}     & \textbf{93.40} & 239 & \textbf{87.64} & 323 & \textbf{82.63} & 762 & \textbf{57.70} & 1358 & \textbf{35.69} & 1788 & \textbf{71.41} & 894.0 \\
\bottomrule
\end{tabular} }

\end{table}

From \Cref{table:ablation}, we can observe that both verifiable signal and conditioned drop measurement are critical for optimal performance. Specifically, $R_\text{EM}$ shows limited accuracy gain but a large efficiency drop, and $R_\text{SA}$ converges to shallow reasoning with fewer tokens and poor accuracy % \footnote{We believe training with $R_\text{SA}$ alone may lead to reward hacking. In \Cref{eq:pplqa}, $R_\text{SA}$ reaches the maximum value 1 only when $ppl(a)\to +\infty$ and $ppl(a|q)=1$. Therefore, the model loses the deep reasoning ability. This is confirmed in \Cref{fig:training}.}
. Failure of these methods indicates that the ground-truth signals are still critical for models to develop reasoning skills and ensure training stability, similar to what's found in~\cite{zhang2025no}

$R_\text{VR}+\alpha R_\text{EM}$ also underperforms SA-GRPO in both accuracy and efficiency, indicating that minimizing entropy is less effective than our approach, which is maximizing conditioned perplexity drop. This is because conditioned perplexity drop provides a more accurate measure of answer quality, as demonstrated in \Cref{sec:case_study}. Moreover, it avoids issues such as overconfidence or entropy collapse, which can hinder exploration~\citep{zhang2025no,cui2025entropy}. Overall, the results suggest ``verifiable signal'' and ``conditioned perplexity drop'' are two crucial components in SAR, without which SA-GRPO won't be able to reach optimal performance in accuracy and efficiency.

% This experiment aims to understand the training process of SA-GRPO by investigating its plots during training. We record three important metrics: correctness reward $R_\text{VR}$, response length, and self-aligned reward $R_\text{SA}$.

% \begin{figure}[htbp]
%     \centering
%     \subcaptionbox{Self-aligned reward\protect\footnotemark.}[0.48\textwidth]{
%         \includegraphics[width=\linewidth]{figures/training_sa.png}
%     }
%     \hfill
%     \subcaptionbox{Response length.}[0.48\textwidth]{
%         \includegraphics[width=\linewidth]{figures/training_len.png}
%     }
%     \subcaptionbox{Base reward.}[0.48\textwidth]{
%         \includegraphics[width=\linewidth]{figures/training_placeholder.png}
%     }
%     \hfill
%     \subcaptionbox{Overall reward.}[0.48\textwidth]{
%         \includegraphics[width=\linewidth]{figures/training_placeholder2.png}
%     }
%     \caption{Training plots for Qwen3-4B.}
%     \label{fig:training}
% \end{figure}

% \footnotetext{While $R_\text{SA}$ isn't used in training the GRPO model, we still calculate and record the values for comparison.}

% From \Cref{fig:training}, we can observe that in the early stage of training, SA-GRPO focuses on increasing accuracy. After the training accuracy plateau, SA-GRPO gradually reduces redundant tokens in reasoning, while maintaining a high accuracy. Moreover, SA-GRPO shows greater potential in long-process optimization, while the baselines' performances converge quickly and fails to further improve.

\subsection{Reasoning Behaviors of SA-GRPO}
\label{sec:reasoning}

Previous work~\citep{gandhi2025cognitive,zeng2025simplerl} has shown that certain reasoning behaviors are critical to effective and deep reasoning. In this section, we investigate four typical behaviors: backtracking, verification, subgoal setting, and enumeration. Specifically, we use GPT-4o to annotate the reasoning behaviors for models trained with different algorithms (see \Cref{app:prompt} for details).

\begin{table}[htbp]
\caption{Frequency of reasoning behaviors. \textbf{B}, \textbf{V}, \textbf{S}, \textbf{E} refer to \textbf{B}acktracking, \textbf{V}erification, \textbf{S}ubgoal setting and \textbf{E}numeration, respectively. Scores are in percentages. The base model is Qwen3-1.7B. Unlike length-based methods, \textbf{SA-GRPO maintains a high usage of reasoning behaviors.}}
\label{table:reasoning}
\small
\centering
\renewcommand{\arraystretch}{1.1}
\setlength{\tabcolsep}{4pt} % Reduces space to 4pt
\begin{tabular}{c|cccc|cccc|cccc|c}
\toprule
       Dataset   & \multicolumn{4}{c}{MATH}                         & \multicolumn{4}{c}{NuminaMath}                          & \multicolumn{4}{c}{AIME}                    &  \multicolumn{1}{c}{\multirow{2}{*}{Average}} \\
Behavior  & \textbf{B} & \textbf{V} & \textbf{S} & \multicolumn{1}{c}{\textbf{E}} & \multicolumn{1}{c}{\textbf{B}} & \textbf{V} & \textbf{S} & \multicolumn{1}{c}{\textbf{E}} & \multicolumn{1}{c}{\textbf{B}} & \textbf{V} & \textbf{S} & \multicolumn{1}{c}{\textbf{E}} & \\ \hline

Base    & 0.8 & 24.2 & 90.8 & 12.2 & 5.4 & 27.0 & 93.4 & 34.8 & 4.4 & 36.2 & 93.2 & 37.4 & 38.0 \\
GRPO      & 1.0 & 29.2 & 95.4 & 14.6 & 4.6 & 38.8 & 97.8 & 44.2 & 5.0 & 42.2 & 98.2 & 49.0 & 43.4 \\
GRPO-O1   & 1.2 & 26.6 & 90.4 & 14.6 & 4.8 & 37.0 & 93.2 & 39.6 & 4.2 & 43.4 & 96.0 & 42.8 & 40.6 \\
GRPO-ER   & 1.4 & 26.0 & 89.8 & 13.4 & 6.2 & 37.0 & 90.4 & 40.4 & 5.8 & 44.2 & 96.2 & 47.6 & 41.2 \\
\rowcolor{mypurple!15}
\textbf{SA-GRPO}  & 0.8 & 29.2 & 93.4 & 13.8 & 7.0 & 36.0 & 93.8 & 43.8 & 8.8 & 47.0 & 97.0 & 49.6 & 43.0 \\
\bottomrule
\end{tabular}
\end{table}

From \Cref{table:reasoning}, we can observe that GRPO-O1/ER exhibit fewer reasoning behaviors compared to GRPO, as reasoning behaviors require additional tokens which are penalized by length-based reward functions. However, \textbf{SA-GRPO maintains almost the same frequency of reasoning behaviors with GRPO}, notably with 30\% fewer tokens than GRPO. This stems from the self-aligned reward's content-aware feature, showcasing its ability to accurately distinguish useful reasoning behaviors from unnecessary content, guiding model behaviors in a more accurate and unbiased manner.

 \begin{wraptable}[11]{r}{0.6\textwidth}
 \vspace{-1em}
\caption{Comparison of training time. We report GPU hours (Training time $\times$ GPU count) of training Qwen3-4B for the first 200 steps.}
\label{table:time}
\small
\centering
\renewcommand{\arraystretch}{1.1}
\begin{tabular}{lcccc}
\toprule
Method    & \textbf{Rollout} & \textbf{Update} & \textbf{Total} \\\hline
% PPO       & 28.92  & 4.40  & 33.32 \\
GRPO      & 32.95  & 15.13 & 48.08 \\
GRPO-O1   & 30.36  & 15.92 & 46.28 \\
GRPO-ER   & 30.92  & 15.24 & 46.16 \\
\textbf{SA-GRPO}      & 31.44  & 15.20 & 46.64 \\
\bottomrule
\end{tabular}

\end{wraptable}

\subsection{Training Cost of self-aligned reward}
\label{sec:time}
 We report the training cost for SA-GRPO in \Cref{table:time}, from which we can find that SA-GRPO doesn't introduce burdens in ``Update'' phase compared to vanilla GRPO. The only additional computation in SAR is $ppl(a)$ ($ppl(a|q)$ is already calculated in GRPO, for KL penalty and importance sampling), which only requires a forward pass, making it highly efficient. Additionally, SA-GRPO even takes less time in the ``Rollout'' phase due to reduced answer lengths. This suggests that enhancing reasoning efficiency is not only valuable during inference -- it can also accelerate RL training and maximize the gain of reasoning capability under limited computation.
\section{Conclusion}

In this work, we propose \textbf{Self-Aligned Reward (SAR)}, an internal perplexity-based signal evaluating the answer's relevancy with the query, enabling fine-grained supervision beyond binary correctness. Through comprehensive experiments on 4 base models and 7 benchmarks, we demonstrated that SAR enables reinforcement learning to achieve consistent gains of up to 4\% in accuracy while reducing response length and computational cost by 30\%. Moreover, SAR exhibits a favorable accuracy–efficiency balance compared with length-based baselines, offering a fine-grained and content-aware reward signal that complements verifiable correctness. Our analysis further shows that SAR generalizes robustly to out-of-domain tasks and preserves advanced reasoning behaviors, underscoring its broad applicability. Our findings highlight the significance of incorporating intrinsic fine-grained model self-assessments into the RL framework.

\newpage
\section*{Reproducibility Statement}

We ensure the reproducibility of the paper from the following aspects:

\textbullet \hspace{1pt} \textbf{Dataset}: In \Cref{app:dataset}, we describe the datasets and their preprocessing methods. All datasets are open-sourced.

\textbullet \hspace{1pt} \textbf{Method}: Our core contribution is self-aligned reward (SAR), a simple yet effective internal signal. We present the formula and explanations of SAR in \Cref{method:sapo}, and show the core code for implementing SAR in \Cref{app:implementation}. In addition, formulations and implementations of the baselines are also presented. Our code is based on VERL, an open-sourced and widely adopted RL framework.

\textbullet \hspace{1pt} \textbf{Training}: Training configurations are presented in \Cref{app:training}. Readers can reproduce the exact training results following these settings.

\textbullet \hspace{1pt} \textbf{Prompting}: The paper's analytical experiments involve annotating with an LLM. Prompts for such annotation are presented in \Cref{app:prompt}.

\section*{LLM Usage Statement}

Large language models were not used in the writing of this paper, except for the sample responses from trained LLM models for illustrative purposes. All written content and experimental code were generated solely by the authors.

\bibliography{iclr2026_conference}
\bibliographystyle{iclr2026_conference}

\newpage
\appendix
\renewcommand\thesection{\Alph{section}}
\renewcommand\thesubsection{\thesection.\arabic{subsection}}
\renewcommand\thesubsubsection{\thesubsection.\arabic{subsubsection}}

% \section*{Appendix Contents}

% \begin{enumerate}
%   \item[A] \textbf{Datasets}
%     \begin{enumerate}
%       \item[A.1] Mathematical Reasoning Benchmarks
%       \item[A.2] Logical Reasoning Benchmarks
%       \item[A.3] Evaluation Details
%     \end{enumerate}

%   \item[B] \textbf{Implementations and Parameter Settings}
%     \begin{enumerate}
%       \item[B.1] Base Models
%       \item[B.2] Implementation Framework
%       \item[B.2] SFT, PPO and GRPO
%       \item[B.3] GRPO (O1) and GRPO (ER)
%       \item[B.4] SA-GRPO
%     \end{enumerate}

%   \item[C] \textbf{Additional Implementation Details}
%     \begin{enumerate}
%       \item[C.1] SA-GRPO Implementation
%       \item[C.2] Perplexity Estimation for SA-GRPO
%       \item[C.3] SA-GRPO Overall Framework
%       \item[C.4] Inference Details
%       \item[C.5] Scaling Experiment Setup
%     \end{enumerate}

%   \item[D] \textbf{Additional Analysis and Theory}
%     \begin{enumerate}
%       \item[D.1] GRPO gradient analysis
%       \item[D.2] Theoretical analysis of SA-GRPO
%         \begin{enumerate}
%           \item[D.2.1] SA-GRPO Reward Function
%           \item[D.2.2] SA-GRPO Optimization Trajectories
%         \end{enumerate}
%     \end{enumerate}

%   \item[E] \textbf{Additional Results and Case Study}

% \end{enumerate}

\section{Datasets}
\label{app:dataset}
We evaluate our approach on five mathematical reasoning datasets with varying difficulty levels, and two logical reasoning datasets. We list the datasets and provide details below.

\subsection{Mathematical Reasoning Benchmarks}

\begin{enumerate}
    \item \textbf{GSM8k} \citep{cobbe2021training}: A dataset of grade school math word problems.
    \item \textbf{MATH} \citep{hendrycks2021measuring}: A challenging dataset of mathematics problems covering various topics.
    \item \textbf{NuminaMath 1.5} \citep{li2024numinamath}: A comprehensive dataset containing 860k pairs of competition math problems and solutions. We select a subset from the Open-r1 project\footnote{\url{https://huggingface.co/datasets/open-r1/OpenR1-Math-220k}}.
    \item \textbf{GSM-symbolic} \citep{mirzadeh2024gsm}: A dataset of GSM8k-style problems with different numbers. This dataset is free from data contamination, making it suitable to evaluate generalization ability.
    \item \textbf{AIME 1983-2024} \citep{veeraboina_aime_2024}\footnote{\url{https://www.kaggle.com/datasets/hemishveeraboina/aime-problem-set-1983-2024}}: A collection of problems from the American Invitational Mathematics Examination spanning over four decades. Requires complex reasoning.
\end{enumerate}

To ensure parsing correctness, we only select questions where the answer is a single integer or fraction number, and remove questions involving geometric plots or hyperlinks. During training, we combined the training splits of GSM8k, MATH, and NuminaMath. GSM-symbolic and AIME were used exclusively for evaluation to test out-of-distribution generalization.

\subsection{Logical Reasoning Benchmarks}
To evaluate the generalization capabilities of our approach beyond mathematical reasoning, we used the following logical reasoning benchmarks:

\begin{enumerate}
    \item \textbf{ProntoQA} \citep{saparov2022language}: A dataset evaluating multi-step syllogistic reasoning.
    \item \textbf{LogicBench} \citep{parmar2024logicbench}: A comprehensive benchmark for evaluating logical reasoning capabilities of language models across various logical relationships.
\end{enumerate}

These datasets were used only for evaluation purposes and were not part of the training process. Both datasets are in the form of multiple-choice questions, where ProntoQA has 2 choices and LogicBench has 4 for each question.

 \Cref{table:datasets} shows the statistics of all datasets used.

\subsection{Evaluation Details}
We construct a unified prompt format for all questions:

\begin{center}
\begin{tcolorbox}[promptstyle]
\textbf{System Prompt}: You are a reasoning expert assistant. Given a question, you will use your reasoning skills to solve the problem.

\textbf{User Prompt}: \textbf{[Question]}

Please explain your reasoning process before providing an answer.
\end{tcolorbox}
\end{center}

During evaluation, we parse the last integer or fraction number in the model's output and compare it with the ground truth using \texttt{math\_verify} package. We don't introduce a strict answer format, as previous work~\citep{zeng2025simplerl} suggest it may hinder exploration.

\begin{table}[t]
\caption{Sizes of datasets.}
\label{table:datasets}
\small
\centering
\renewcommand{\arraystretch}{1.1}
\setlength{\tabcolsep}{4pt} % Reduces space to 4pt
\begin{tabular}{lccccccc}
\toprule
Dataset     & GSM8k & MATH & NuminaMath & GSM-sym & AIME & ProntoQA & LogicBench \\\hline
\# Train & 7473  & 5654 & 10000      & -       & -    & -        & -          \\
\# Eval     & 1319  & 3742 & 2742       & 5000    & 933  & 500      & 500       \\\bottomrule
\end{tabular}
\end{table}

\section{Implementation Details}
\label{app:implementation}

\subsection{Implementation of SAR}
We implemented all baselines, as well as our approach building on the VERL open-source framework~\cite{sheng2025hybridflow}. The VERL framework provides standard implementations of PPO and GRPO. To implement our approach, we need to calculate the perplexity values for the rollouts given the query ($\operatorname{ppl}(a|q)$) as well as the perplexity of standalone rollouts ($\operatorname{ppl}(a)$).

In most RL algorithms, calculating log probabilities for rollouts are necessary to apply KL penalty. These log-probs can be directly applied to calculate $\operatorname{ppl}(a|q)$ without extra cost:

\begin{lstlisting}[style=mypython]
def compute_ppl(log_probs, mask):
    '''
    log_probs: Tensor[batch_size, seq_len]
    mask: Boolean Tensor[batch_size, seq_len], positions of the answer part are 1; positions of query and padding tokens are 0.
    '''
    sum_log_probs = (log_probs * mask).sum(dim=1)
    lengths = mask.sum(dim=1)
    avg_log_probs = sum_log_probs / lengths
    return torch.exp(-avg_log_probs)
\end{lstlisting}

The calculation of $\operatorname{ppl}(a)$ consists of two steps. The first step is to construct the sequence without the user prompt (we still keep the system prompt):

\begin{lstlisting}[style=mypython]
def construct_empty_question_data(data, tokenizer, q_max_len):
    '''
    data: DataProto, a standard protocol for data exchange in VERL.
    data contains "input_ids", "attention_mask" and "position_ids", which are Tensors[batch_size, q_max_len+a_max_len]. Query are left-padded and answers are right-padded, which means the answer always begin at index q_max_len.
    '''
    empty_q = f"System:{sys_prompt}\nUser:\nAssistant:"
    empty_q_tokens = tokenizer(empty_q, padding="left", max_length=q_max_len)

    batch_size = data["input_ids"].shape[0]
    for key in ["input_ids", "attention_mask", "position_ids"]:
        data[key][:, :q_max_len] = empty_q_tokens[key].repeat(batch_size, 1)

    last_query_pos = data["position_ids"][:, q_max_len - 1]
    first_answer_pos = data["position_ids"][:, q_max_len] # the first answer token
    gap = first_answer_pos - last_query_pos - 1 # this gap should be zero
    data["position_ids"][:, q_max_len:] -= gap.unsqueeze(-1) # Shift the answer positions backward by this gap so they are continuous

    return data
\end{lstlisting}

The second step is to calculate log probabilities for the ``new'' sequences. Combining them, we implement the self-aligned reward:
\begin{lstlisting}[style=mypython]
def R_SA(data, tokenizer, q_max_len, policy):
    ppl_qa = compute_ppl(data["log_probs"], data["response_mask"])

    empty_q_data = construct_empty_question_data(data, tokenizer, q_max_len)
    empty_q_log_probs = policy.calc_log_probs(empty_q_data)
    ppl_a = compute_ppl(empty_q_log_probs, data["response_mask"])

    return max((ppl_a - ppl_qa) / ppl_a, -1)
    
\end{lstlisting}

In conclusion, the self-aligned reward calculation is fully compatible with VERL's existing optimizations, making it highly efficient for training large models.

\subsection{Implementation of GRPO-O1 and GRPO-ER}
\label{method:baseline}

Signals besides verifiable reward have also been widely studied, where enhancing efficiency is a prevalent direction. In this paper, we consider O1-pruner~\citep{luo2025o1} and Efficient Reasoner (ER)~\citep{arora2025training}, two typical length-based methods as baselines. Since the original works used offline RL algorithms, we made slight changes to the formula in order to adapt to GRPO:

\begin{equation}
\label{eq:O1}
\small
R_\text{GRPO-o1}(q,a_i,gt)=R_\text{VR}+\alpha R_\text{o1}, 
R_\text{o1}= \operatorname{clip}(\frac{\operatorname{mean}[\operatorname{len}(a_{1...N})]-\operatorname{len}(a_i)}{\operatorname{len}(a_i)},-1, 1)
\end{equation}

\vspace{-1.5em}

\begin{equation}
\label{eq:ER}
\small
R_\text{GRPO-ER}(q,a_i,gt)=R_\text{VR}+\alpha R_\text{ER}, 
R_\text{ER}=
\begin{dcases} 2 * \sigma (\frac{\operatorname{mean}[\operatorname{len}(\mathbf{A}_\text{cor})]-\operatorname{len}(a_i)}{\operatorname{std}[\operatorname{len}(\mathbf{A}_\text{cor})]+eps})- 1 & \text{if } R_\text{VR}=1, \\ 
0 & \text{otherwise},
\end{dcases}
\end{equation}

where $\sigma(x)=\frac{1}{1+e^{-x}}$ is the Sigmoid function, and $\mathbf{A}_\text{cor}=\{a | a\in a_{1...N} \land R_\text{VR}(q,a,gt)=1\}$ refers to the set of all correct answers.

% \peixuan{Should we include the comparison of confidence-based methods in thie section?}

We also provide the pseudocode for calculating $R_\text{o1}$ and $R_\text{ER}$ for reference.

\begin{lstlisting}[style=mypython]

def R_O1(length: List[int], correctness: List[bool]):
    avg = statistics.mean(length)
    o1_scores = []
    for len in length:
        x = (avg - len) / len
        x = max(min(x, 1), -1)
        o1_scores.append(x)
    return o1_scores

def R_ER(length: List[int], correctness: List[bool]):
    if correctness.count(True) <= 1:
        return [0] * len(length)
    cor_lens = [len for len, correct in zip(length, correctness) if correct]
    avg = statistics.mean(cor_lens)
    std = statistics.stdev(cor_lens)
    er_scores = []
    for len, cor in zip(length, correctness):
        if cor:
            x = (avg - len) / std
            x = 2 / (1 + math.exp(-x)) - 1
            er_scores.append(x)
        else:
            er_scores.append(0)
    return er_scores
\end{lstlisting}

\subsection{Training Settings}
\label{app:training}

\begin{wraptable}[14]{r}{0.5\textwidth}
\vspace{-2em}
\caption{Training configs for PPO and GRPO.}
\label{table:hparam}
\small
\centering
\renewcommand{\arraystretch}{1.1}
\begin{tabular}{lcc}
\toprule
        Hyperparameter                    & PPO               & GRPO      \\\hline
Actor learning rate                    & \multicolumn{2}{c}{$1e-6$} \\
Critic learning rate                   & $2e-6$       & -         \\
train\_batch\_size\footnotemark          & \multicolumn{2}{c}{128}       \\
mini\_batch\_size           & \multicolumn{2}{c}{64}        \\
micro\_batch\_size          & \multicolumn{2}{c}{16}        \\
Training step                        & \multicolumn{2}{c}{500}       \\
Max response length                     & \multicolumn{2}{c}{4096}      \\
Num of rollouts             & -                 & 8         \\
Rollout temp ($\tau$)       & \multicolumn{2}{c}{1.0}       \\
KL penalty ($\beta$)        & \multicolumn{2}{c}{$1e-3$}     \\
Advantage clip ($\epsilon$) & \multicolumn{2}{c}{0.2}       \\\bottomrule
\end{tabular}
\end{wraptable}

\footnotetext{The three batch sizes in the table are hyperparameters in the VERL framework. train\_batch\_size is the batch size for sampling rollouts, mini\_batch\_size is the batch size to perform policy updates, and micro\_batch\_size is the batch size for rollouts and back-propagation. Theoretically, micro\_batch\_size won't affect the training result, so one can set a smaller or larger value depending on the compute resources.}

Models are trained on 4 NVIDIA H100 80GB GPUs. %  For SFT, we use a batch size of 32 and a constant learning rate of $10^{-5}$. We first sample 8 rollouts per question with $\tau=1$ and select the shortest correct rollout (if there are no correct rollouts for a question, we skip it). Then the models are trained on these rollouts for 1 epoch (training for longer will lead to over-fitting). 
For PPO and GRPO, we list all training hyperparameters in \Cref{table:hparam}. PPO and SA-PPO use the same config; GRPO, GRPO-O1, GRPO-ER and SA-GRPO also use the same config; the only exception is $\alpha$ in the reward function. The default $\alpha$ for GRPO-O1 and GRPO-ER are 0.05, and 0.2 for SA-GRPO. Using a different $\alpha$ will make these algorithms focus more on accuracy or efficiency, as shown in \Cref{fig:contours}. We use vLLM~\citep{kwon2023efficient} framework for inference. Models are evaluated using 1 NVIDIA G100 80GB GPU. By default, the maximum response length is 4096, and greedy decoding is used ($\tau=0$).

\newpage
\section{LLM Annotation Prompts}
\label{app:prompt}

Below is the prompt to categorize responses to 4 types in \Cref{sec:case_study}:

\begin{center}
\begin{tcolorbox}[promptstyle]
\textbf{System Prompt:}
You are an expert reasoner and LLM judge. Given a reasoning problem and an answer, you need to category the answer into one of the following categories:

1: Correct and concise answer. Small mistakes are acceptable.

2: Correct answer, but a bit lengthy, or contains unnecessary steps. Small mistakes are acceptable.

3: Partly correct answers, which makes some mistake and fails to reach the final ground truth.

4: Completely wrong or irrelevant answers, indicating the model doesn't understand the problem.

0: The answer is correct but contains extra content after the answer, like random characters or talking about an irrelevant topic.
\\\\
**Output Format:**

[thought] Provide your thought process on how you identify the reasoning behaviors.[/thought]

[answer] One single number, indicating the type of the answer. [/answer]
\\\\
\textbf{The one-shot example:}

Query: Making a cake requires 1 cup of flour, 1 cup of sugar, and 2 eggs. Suppose you have 2 cups of flour, 3 cups of sugar, and 6 eggs, how many cakes can you make?

Ground truth answer: 2

Answer: I shall first determine how many cakes each ingredient can support. Flour allows 2 cakes (2/1=2), sugar allows 3 (3/1=3), and eggs allow 3 (6/2=3). Therefore, I can make at most 3 cakes. Wait a second, the maximum number actually depends on the the low-resource ingredient. In this case, it's flour, which only supports 2 cakes. The final answer is 2.
\\

[thought] The answer correctly understands the problem and provides a concise solution. It identifies the limiting ingredient (flour) and calculates the maximum number of cakes that can be made based on that. The reasoning is clear and follows a logical sequence. [/thought]

[answer] 1 [/answer]
\end{tcolorbox}
\end{center}

Below is the prompt to identify reasoning behaviors in \Cref{sec:reasoning}:

\begin{center}
\begin{tcolorbox}[promptstyle]
\textbf{System Prompt:}
You are an expert reasoner and LLM judge. Given a reasoning problem and an answer, you need to identify the reasoning behaviors exhibited in the solution. There are four reasoning behaviors that requires identification:

1. **Verification**: This behavior involves systematically checking intermediate results or assumptions to ensure they are correct.

    - Example: "Let's verify this result by checking if the two expressions are always equal."
    
2. **Backtracking**: This behavior occurs when the model explicitly revises its approach after detecting an error or realizing that the current path won't lead to the solution.

    - Example: "The assumption that $a>0$ doesn't work, we need to try something else."
    
3. **Subgoal Setting**: This behavior involves breaking down a complex problem into smaller, more manageable steps.

    - Example: "The first step is to find the range of a and b respectively."
    
4.  **Enumeration**: Solving problems by exhaustively considering multiple cases or possibilities.

    - Example: "After investigating 7 days of a week, only Wedensday and Friday satisfies the conditon."
\\\\
**Guidelines for Identification:**

- It is possible for a single solution to exhibit multiple behaviors or none of them.

- Your annotation should based on the reasoning process, not just the final answer.

- You should only count a behavior if it concretely contributes to the reasoning process.
\\\\
**Output Format:**

[thought] Provide your thought process on how you identify the reasoning behaviors. [/thought]

[answer] Behavior(s) separated by commas, or "None" if no behavior is identified. [/answer]
% \\\\
% \textbf{The one-shot example:}

% Query: Making a cake requires 1 cup of flour, 1 cup of sugar, and 2 eggs. Suppose you have 2 cups of flour, 3 cups of sugar, and 6 eggs, how many cakes can you make?

% Answer: I shall first determine how many cakes each ingredient can support. Flour allows 2 cakes (2/1=2), sugar allows 3 (3/1=3), and eggs allow 3 (6/2=3). Therefore, I can make at most 3 cakes. Wait a second, the maximum number actually depends on the the low-resource ingredient. In this case, it's flour, which only supports 2 cakes. The final answer is 2.
% \\

% [thought] The answer first decomposes the original problem into subproblems, determining how many cakes can be made based on each ingredient. In addition, it corrects its own mistake by pointing out it should focus on low-resource ingredient, which is a typical backtracking behavior.[/thought]

% [answer] Subgoal Setting, Backtracking [/answer]
\end{tcolorbox}
\end{center}

\section{Additional Analysis and Theory}

\subsection{GRPO Gradient Analysis}
\label{sec:grpo_grad}

To understand how GRPO and our SA-GRPO approach optimize model parameters, we analyze the gradient computation. The gradient of the GRPO objective is:

\begin{align}
\nabla_\theta \mathcal{J}_\text{GRPO}(\theta)
&= \nabla_\theta\ \mathbb{E}_{q \sim \mathcal{D},\, \{a_i\}_{i=1}^N \sim \pi_{\theta_\text{old}}(\cdot \mid q)}
\Bigg[
    \frac{1}{N}\sum_{i=1}^{N} \frac{1}{|a_i|}
    \sum_{j=1}^{|a_i|} w_{i,j}(\theta)\, A_i
\Bigg] 
\\
&= \mathbb{E}_{q \sim \mathcal{D},\, \{a_i\}_{i=1}^N \sim \pi_{\theta_\text{old}}(\cdot \mid q)}
\Bigg[
    \frac{1}{N}\sum_{i=1}^{N} A_i \cdot \frac{1}{|a_i|}
    \sum_{j=1}^{|a_i|}
        r_{i,j}\,
        \nabla_\theta \log \pi_\theta(a_{i,j} \mid q, a_{i,<j})
\Bigg]
\end{align}

\noindent
where 
$$
r_{i,j} = \frac{\pi_\theta(a_{i,j}\mid q, a_{i,<j})}
               {\pi_{\theta_\text{old}}(a_{i,j}\mid q, a_{i,<j})}, 
\quad
w_{i,j}(\theta) = r_{i,j}.
$$

This gradient formulation reveals that GRPO updates model parameters by adjusting the likelihood of generating each token in the rollout responses, weighted by the importance sampling ratio $r_{i,j}$ and the normalized advantage $A_i$. The advantage is calculated based solely on the verifiable reward or other explicit reward functions, without considering the intrinsic quality or relevance of responses.

In standard GRPO, the advantage value $A_i$ is the same for all tokens within a single answer, computed as:
\begin{equation}
A_i = R(q,a_i) - \text{mean}[R(q,a_{1...N})]
\end{equation}

For correct answers, the advantage is positive, pushing the model to increase the probability of generating such answers. For incorrect answers, the advantage is negative, steering the model away from these outputs.

However, this approach treats all correct answers equally, regardless of their efficiency, relevance, or quality of reasoning. Similarly, it gives all incorrect answers the same negative feedback, missing opportunities to reinforce partially correct reasoning paths or penalize completely irrelevant outputs differently.

\subsection{Theoretical Analysis of SAR}
\label{sec:sapo_theory}

\subsubsection{Self-aligned Reward Function}
\label{sec:sapo_reward}

The key innovation, self-aligned reward $R_\text{SA}$, can be derived from the perplexity measures:

\begin{align}
R_\text{SA} &= \max\left(\frac{\operatorname{ppl}(a) - \operatorname{ppl}(a|q)}{\operatorname{ppl}(a)}, -1\right) \\
&= 1 - \min\left(\frac{\operatorname{ppl}(a|q)}{\operatorname{ppl}(a)}, 2\right)
\end{align}

Expanding with the definitions of perplexity:
\begin{align}
R_\text{SA} &= 1 - \min\left(\frac{e^{-\frac{1}{|a|}\sum_{j=1}^{|a|} \log(P(a_j|q,a_{1...j-1}))}}{e^{-\frac{1}{|a|}\sum_{j=1}^{|a|} \log(P(a_j|a_{1...j-1}))}}, 2\right) \\
&= 1 - \min\left(e^{-\frac{1}{|a|}\sum_{j=1}^{|a|} \left[\log(P(a_j|q,a_{1...j-1})) - \log(P(a_j|a_{1...j-1}))\right]}, 2\right) \\
&= 1 - \min\left(e^{-\frac{1}{|a|}\sum_{j=1}^{|a|} \log\left(\frac{P(a_j|q,a_{1...j-1})}{P(a_j|a_{1...j-1})}\right)}, 2\right)
\end{align}

Defining the token-level contribution $v(a_j) = \log\left(\frac{P(a_j|q,a_{1...j-1})}{P(a_j|a_{1...j-1})}\right)$, we get:
\begin{equation}
R_\text{SA} = 1 - \min\left(e^{-\frac{1}{|a|}\sum_{j=1}^{|a|} v(a_j)}, 2\right)
\end{equation}

This formulation reveals that $R_\text{SA}$ measures the geometric mean of the likelihood ratio between generating tokens conditioned on the question versus generating them without the question context. Each token $a_j$ contributes $v(a_j)$ to the overall reward.

A positive $v(a_j)$ indicates that token $a_j$ is more likely to be generated when conditioned on the question, suggesting that it leverages information from the query. Conversely, a negative $v(a_j)$ suggests that the token is less likely when conditioned on the question, indicating potential irrelevance or redundancy.

\subsubsection{SA-GRPO Optimization Trajectories}
\label{sec:sapo_opt}

The combined reward $R_\text{SA-GRPO}(q,a_i,gt) = R_\text{VR} + \alpha R_\text{SA}$ leads to a modified advantage calculation:

\begin{equation}
A_i^{\text{SA-GRPO}} = (R_\text{VR}(q,a_i,gt) + \alpha R_\text{SA}(q,a_i)) - \text{mean}[(R_\text{VR}(q,a_{1...N},gt) + \alpha R_\text{SA}(q,a_{1...N}))]
\end{equation}

During optimization, SA-GRPO's gradient updates follow three key paths:

1. \textbf{Correctness Optimization}: Through $R_\text{VR}$, SA-GRPO increases the likelihood of generating outputs that contain correct answers, similar to standard GRPO.

2. \textbf{Self-Alignment Optimization}: Through $R_\text{SA}$, SA-GRPO encourages:
   - Higher probability for tokens that effectively utilize question information (positive $v(a_j)$)
   - Lower probability for tokens that are redundant or irrelevant (negative $v(a_j)$)
   - Concise reasoning by penalizing unnecessary repetition, since repeated information yields low $v(a_j)$ values as it becomes predictable from previous tokens

3. \textbf{Memorization Penalty}: SAR naturally penalizes memorized answers. If an answer is memorized, $\operatorname{ppl}(a)$ will be low due to the model's strong prior on the memorized sequence, while the relative drop in perplexity when conditioned on the question would be minimal, resulting in a low or negative $R_\text{SA}$.

\begin{figure}[htbp]
    \centering
    \subcaptionbox{Verifiable reward.}[0.32\textwidth]{
        \includegraphics[width=\linewidth]{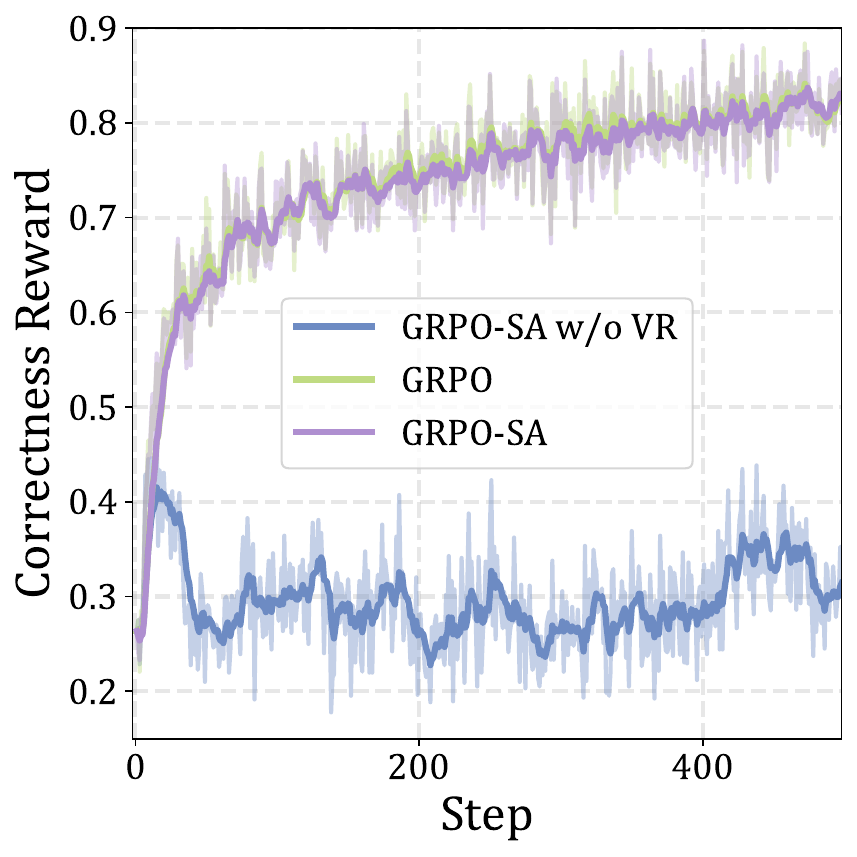}
    }
    \hfill
    \subcaptionbox{Self-aligned reward\protect\footnotemark.}[0.32\textwidth]{
        \includegraphics[width=\linewidth]{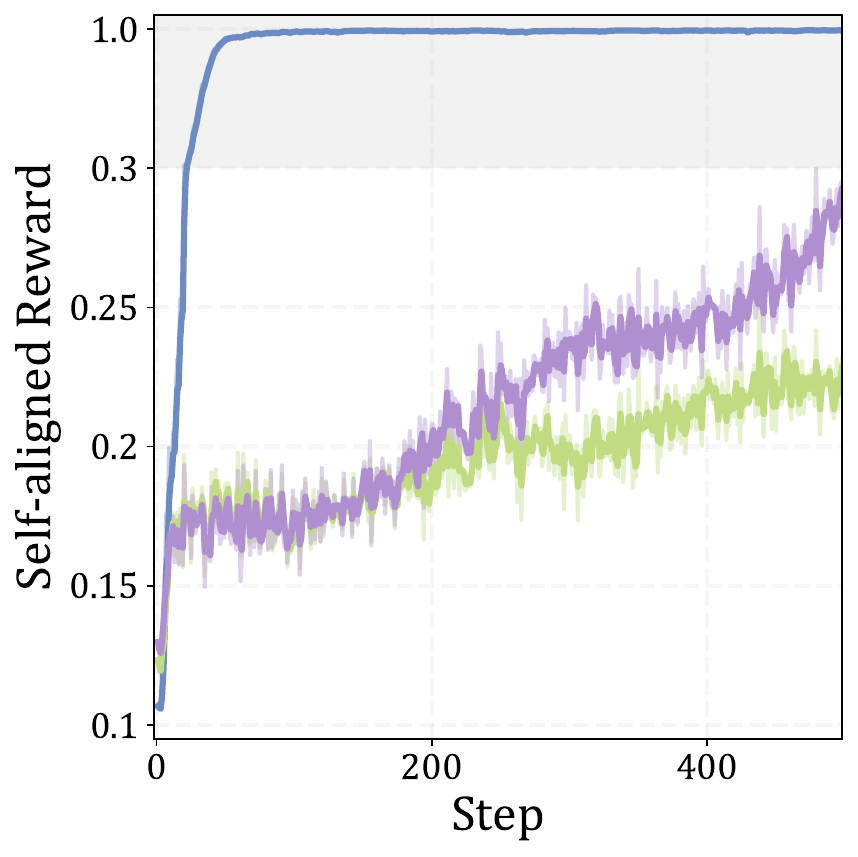}
    }
    \hfill
    \subcaptionbox{Average answer length.}[0.32\textwidth]{
        \includegraphics[width=\linewidth]{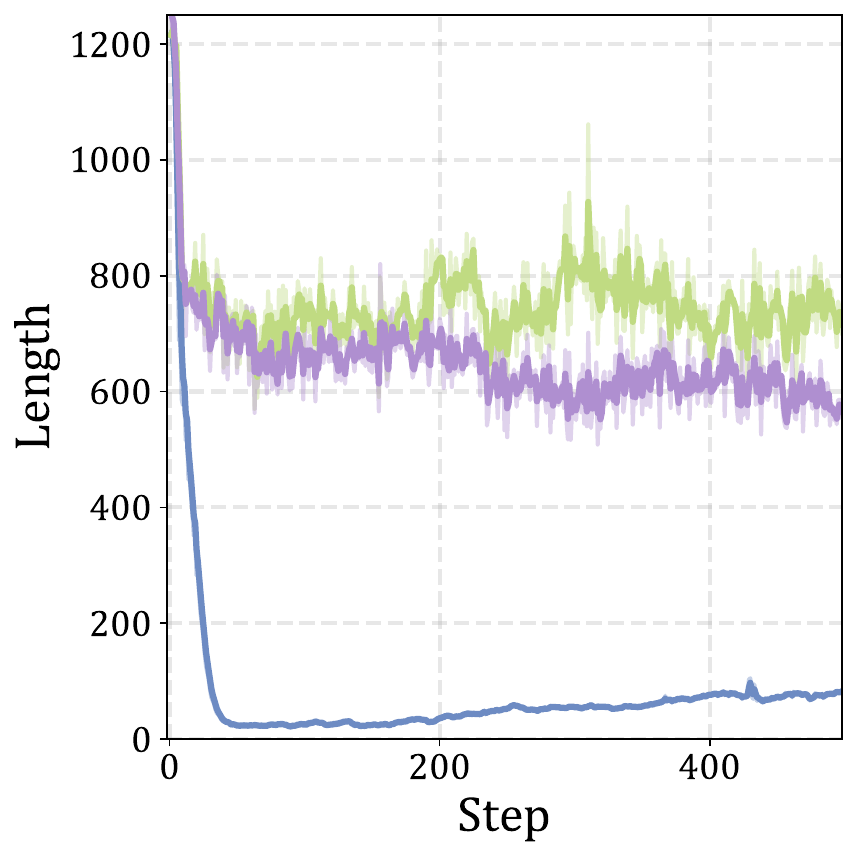}
    }
    \caption{Training plots for Qwen3-4B.}
    \label{fig:training}
\end{figure}

These optimization paths allow SA-GRPO to simultaneously improve accuracy and efficiency without requiring separate reward models or human preference data. The $\alpha$ hyperparameter controls the balance between correctness and self-alignment objectives, enabling flexible tuning for different accuracy-efficiency trade-offs. As training progresses, we observe from \cref{fig:training}:

% Theoretically, SA-GRPO's gradient updates will converge to a policy that generates answers that are both correct and optimally self-aligned—containing only tokens that effectively utilize information from the question while avoiding unnecessary repetition or irrelevant content.

% The trajectory of SA-GRPO optimization can be understood through the lens of token-level contribution analysis. 

1. \textbf{Early Training Phase}: During early iterations, SA-GRPO primarily optimizes for correctness, as $R_\text{VR}$ provides the strongest gradient signal. This establishes a foundation of accurate reasoning.

2. \textbf{Mid Training Phase}: Once the model achieves reasonable accuracy, the self-alignment reward $R_\text{SA}$ becomes more influential. The optimization begins to focus on improving the efficiency of correct responses by:
   - Removing tokens with low $v(a_j)$ values (those that don't effectively leverage question information)
   - Preserving tokens with high $v(a_j)$ values (those that directly address the question)

3. \textbf{Late Training Phase}: In the later stages, SA-GRPO fine-tunes the balance between accuracy and efficiency. The combined reward creates a Pareto frontier where further improvements in efficiency come at diminishing costs to accuracy.

Unlike efficiency-focused methods like O1-pruner or Efficient Reasoner, which directly reward shorter responses regardless of content quality, SA-GRPO's optimization is content-aware. It selectively preserves tokens that contribute meaningful information relative to the question, while removing those that don't. In contrast, length-based rewards may inadvertently remove important reasoning steps if they blindly optimize for shorter responses.

% This content-aware optimization explains why SA-GRPO can simultaneously improve both accuracy and efficiency—it removes only the unnecessary tokens while preserving or even enhancing the essential reasoning steps. 
The theoretical convergence of SA-GRPO can be expressed as finding the optimal policy $\pi_{\theta}^*$ that maximizes:

\begin{equation}
\pi_{\theta}^* = \arg\max_{\pi_{\theta}} \mathbb{E}_{q \sim \mathcal{D}, a \sim \pi_{\theta}(\cdot|q)} \left[ R_\text{VR}(q, a, gt) + \alpha \max\left(\frac{\operatorname{ppl}(a) - \operatorname{ppl}(a|q)}{\operatorname{ppl}(a)}, -1\right) \right]
\end{equation}

At this optimal policy, each generated token in the response contributes maximally to either obtaining the correct answer or efficiently utilizing information from the question, with minimal redundancy or irrelevance.

In practice, this theoretical optimum must balance against the KL divergence penalty that prevents the model from straying too far from the reference policy, ensuring that the learned improvements remain grounded in the model's original capabilities.

\section{Additional Experiments}

\subsection{Training Trajectories}
\label{app:plots}

In this section, we show the training trajectories of GRPO, SA-GRPO and SA-GRPO without verifiable reward (a setting discussed in \Cref{sec:ablation}). We report the verifiable reward, self-aligned reward, and average response length at each step.

\footnotetext{While $R_\text{SA}$ isn't used in training the GRPO model, we still calculate and record the values for comparison.}

\subsection{Self-aligned Reward on Vision Language Models}

\begin{table}[t]
\caption{SAR results on vision-language models.}
\vspace{0.4em}
\label{table:result_vlm}
\small
\centering
\renewcommand{\arraystretch}{1.05}
\setlength{\tabcolsep}{4pt} % Reduces space to 3pt

\resizebox{1.0\textwidth}{!}{%
\begin{tabular}{lcccccccccc|cc}
\toprule
\multirow{2}{*}{\textbf{Setting}} & \multicolumn{2}{c}{GSM8k} & \multicolumn{2}{c}{GSM-sym} & \multicolumn{2}{c}{MATH} & \multicolumn{2}{c}{NuminaMath} & \multicolumn{2}{c}{AIME} & \multicolumn{2}{c}{\textbf{Average}} \\
           & acc & len & acc & len & acc & len & acc & len & acc & len & acc & len \\ 
\midrule \midrule
\rowcolor{gray!15}
\textbf{Qwen2-2B-VL} & 14.33 & 1573 & 5.98 &1873  & 14.94 & 15.84 & 3.65 & 2411 & 0.54 & 2404 & 7.89 & 1969.0 \\
~~~+GRPO      & 60.96 & 150 & 27.72 & 209 & 25.25 & 321 & 9.19 & 512 & 0.86 & 561 & 24.80 & 350.6 \\
\rowcolor{gray!10}
~~~+GRPO-O1   & 58.38 & 115 & 25.44 & 164 & 23.20 & 79 & 9.99 & 22 & 0.32 & 17 & 23.47 & 79.4 \\
~~~+GRPO-ER   & 58.30 & 122 & 26.52 & 179 & 23.73 & 212 & 8.32 & 330 & 0.43 & 383 & 23.46 & 245.2 \\
\rowcolor{mypurple!10}
~~~+SA-GRPO      & 59.59 & 121 & 27.66 & 174 & 21.22 & 84 & 9.52 & 56 & 1.29 & 57 & 23.86 & 98.4 \\
\noalign{\vskip 2pt}
\rowcolor{gray!15}
\textbf{Gemma3-4B}   & 74.75 & 264 & 57.40 & 373 & 68.33 & 837 & 35.81 & 1370 & 18.76 & 1843 & 51.01 & 937.4 \\
~~~+GRPO      & 89.01 & 409 & 78.68 & 644 & 76.54 & 1177 & 41.9 & 2012 & 18.01 & 2357 & 60.83 & 1355.8 \\
\rowcolor{gray!10}
~~~+GRPO-O1   & 87.87 & 292 & 75.02 & 464 & 76.67 & 834 & 41.68 & 1435 & 18.54 & 1814 &  59.95& 967.8 \\
~~~+GRPO-ER   & 88.17 & 338 & 73.36 & 623 & 75.73 & 979 &41.94  & 1817 & 17.36 &2268  & 59.31 & 1205.0 \\
\rowcolor{mypurple!10}
~~~+SA-GRPO      & 89.16 & 407 & 79.04 & 654 & 75.55 & 1130 & 40.48 & 1920 & 18.11 & 2379 & 60.47 & 1298.0 \\
\bottomrule
\end{tabular} }

\end{table}

In this section, we extend SA-GRPO to vision language models.

\Cref{table:result_vlm} summarizes the performance of Qwen2-2B-VL and Gemma3-4B across five reasoning benchmarks under different optimization strategies. Consistent with prior findings, SA-GRPO yields strong gains for purely text-based reasoning—most notably in GSM8k, MATH, and Numina—where its self-aligned reward mechanism effectively tailors outputs to the specific query. For Gemma3-4B, SA-GRPO achieves the highest average accuracy (72.64\%), exceeding the best GRPO variant by over one percentage point, while maintaining competitive output lengths. These improvements suggest that SA-GRPO’s reward shaping promotes concise, context-aware reasoning paths in textual domains.

However, the results also reveal a notable limitation: SA-GRPO’s advantages do not transfer as strongly to vision-language models such as Qwen2-2B-VL. While modest gains are observed in certain benchmarks, its performance is less consistent compared to GRPO, particularly in tasks like GSM-sym and AIME where visual interpretation is coupled with reasoning. We hypothesize that SA-GRPO’s self-alignment mechanism, optimized for semantic relevancy in text, is less effective when the query interpretation depends heavily on multimodal fusion. In such cases, reward signals based solely on linguistic alignment may fail to capture errors introduced in the visual grounding stage, leading to weaker overall gains.

These findings suggest that while SA-GRPO is well-suited for text-centric reasoning, its application to multimodal settings may require integrating visual-aware reward components. Future work could explore hybrid reward functions that jointly evaluate semantic correctness and perceptual grounding, enabling SA-GRPO to extend its benefits to vision-language reasoning tasks.

\subsection{Effect of Removing the KL Coefficient}

\begin{table}[hbtp]
\caption{Effect of removing the KL penalty. The base model is Qwen3-1.7B in this experiment.}
\vspace{0.4em}
\label{table:nokl}
\small
\centering
\renewcommand{\arraystretch}{1.05}
\setlength{\tabcolsep}{4pt} % Reduces space to 3pt

\resizebox{1.0\textwidth}{!}{%
\begin{tabular}{lcccccccccc|cc}
\toprule
\multirow{2}{*}{Setting} & \multicolumn{2}{c}{GSM8k} & \multicolumn{2}{c}{GSM-sym} & \multicolumn{2}{c}{MATH} & \multicolumn{2}{c}{NuminaMath} & \multicolumn{2}{c}{AIME} & \multicolumn{2}{c}{Average} \\
           & acc        & len        & acc         & len        & acc          & len         & acc          & len          & acc         & \multicolumn{1}{c}{len}        & \multicolumn{1}{c}{acc}     & len     \\ 
\midrule \midrule
\rowcolor{gray!15}
Base & 69.22 & 281 & 46.76 & 365 & 56.89 & 700 & 24.58 & 1242 & 9.97 & 1539 & 41.48 & 825.4 \\
GRPO      & 84.53 & 335 & 67.66 & 413 & 67.31 & 697 & 33.44 & 1091 & 13.93 & 1278 & 53.37 & 762.8 \\
\rowcolor{gray!10}
GRPO (no KL)      & 84.69 & 341 & 69.72 & 430 & 68.31 & 685 & 33.55 & 1121 & 13.61 & 1320 & 53.97 & 779.0 \\
GRPO-O1   & 84.15 & 285 & 65.78 & 363 & 66.51 & 528 & 33.15 & 775 & 13.61 & 910 & 52.64 & 572.2 \\
\rowcolor{gray!10}
GRPO-O1 (no KL)   & 83.93 & 275 & 66.40 & 356 & 67.13 & 578 & 32.46 & 875 & 14.15 & 1086 & 52.82 & 634 \\
SA-GRPO      & 85.51 & 267 & 67.66 & 346 & 67.96 & 564 & 36.03 & 841 & 13.50 & 992 & 54.13 & 602.0 \\
\rowcolor{gray!10}
SA-GRPO (no KL)      & 85.75 & 246 & 67.62 & 322 & 66.73 & 511 & 34.06 & 788 & 12.43 & 933 & 53.31 & 560 \\
\bottomrule
\end{tabular}
 }

\end{table}

In \Cref{eq:grpo}, the KL penalty term is used to control the divergence between the reference model and the actor policy, which is critical to preserving the base model's capabilities. We investigate the effect of removing the KL Coefficient (i.e. setting $\beta=0$ in \Cref{eq:grpo}) in this section.

From \Cref{table:nokl}, we observe that for all RL algorithms, the performance gap between using and not using KL is relatively small. 
This indicates that training remains stable and does not diverge significantly from the base model. 
It also suggests that the self-aligned signal is sufficiently consistent and well-aligned to guide learning even without strong distributional anchoring.

Notably, removing the KL penalty produces opposite effects for GRPO/GRPO-O1 and SA-GRPO. 
For GRPO and GRPO-O1, it increases accuracy but reduces efficiency, while for SA-GRPO, it decreases accuracy but improves efficiency. 
Since the verifiable reward $R_{\text{VR}}$ is directly linked to accuracy and the self-aligned reward $R_{\text{SA}}$ is not, this pattern suggests that $R_{\text{VR}}$ dominates in GRPO-O1 (reward $= R_{\text{VR}} + \alpha R_{\text{o1}}$), whereas the self-aligned reward $R_{\text{SA}}$ dominates in SA-GRPO (reward $= R_{\text{VR}} + \alpha R_{\text{SA}}$). 
Across all variants, there's a consistent trade-off between accuracy and efficiency: moving toward higher accuracy tends to reduce efficiency, and vice versa.

\newpage
\section{Cases}

\subsection{Different Answer Types and Advantage Values}

\begin{table}[hbtp]
\caption{An example for different types of answers and their advantage values.}
\label{table:category_case}
\small
\centering
\renewcommand{\arraystretch}{1.1}
\setlength{\tabcolsep}{4pt} % Reduces space between columns
\resizebox{1.0 \textwidth}{!}{%
\begin{tabularx}{\textwidth}{>{\raggedright\arraybackslash}p{0.51\textwidth} >{\raggedright\arraybackslash}p{0.075\textwidth}>{\raggedright\arraybackslash}p{0.075\textwidth}>{\raggedright\arraybackslash}p{0.075\textwidth}>{\raggedright\arraybackslash}p{0.075\textwidth}>{\raggedright\arraybackslash}p{0.075\textwidth}}
\toprule
\multicolumn{6}{>{\raggedright\arraybackslash}X}{
\textbf{Query:} Janet's ducks lay 16 eggs per day. She eats three for breakfast every morning and bakes muffins for her friends every day with four. She sells the remainder at the farmers' market daily for \$2 per fresh duck egg. How much in dollars does she make every day at the farmers' market?
} \\\hline
\textbf{Answer Type and Example} & \textbf{Length} & $A_\text{GRPO}$ & $A_\text{GRPO-o1}$ & $A_\text{GRPO-ER}$ & $\ A_\text{SA-GRPO}$\ \ (Ours)\\
\hline
\textbf{(1) Correct and Concise:} We can break this down step by step: Janet's ducks lay 16 eggs per day. She eats 3 eggs for breakfast every morning. So, 16 - 3 = 13 eggs remain after Janet eats breakfast. She bakes 4 eggs in muffins every day for her friends. So, we need to subtract these as well: 13 - 4 = 9 eggs. Hence, Janet sells 9 eggs to the farmers' market daily. She sells each egg at \$2. Therefore Janet's daily revenue is 9 eggs * \$2 = \$18. So, Janet makes \$18 per day at the farmers' market. & 143 & 0.5 & 0.64 & 0.63 & 0.99 \\
\textbf{(2) Correct but redundant:} To answer the question, we need to first calculate how many eggs Janet uses for breakfast and for baking daily. Janet eats 3 eggs for breakfast every morning. Additionally, she bakes muffins with 4 eggs daily. Let's sum the total number of eggs used for both breakfast and baking: 3 (breakfast) + 4 (baking) = 7 eggs total used daily. Now that we know she uses 7 eggs daily, we can determine how many fresh eggs she has left after accounting for her usage. Since her ducks lay 16 eggs per day, we subtract her total egg usage from the daily egg production: 16 (total eggs produced) - 7 (total eggs used) = 9 fresh eggs remaining. Finally, to find out how much money Janet makes selling the fresh eggs, we multiply the number of remaining fresh eggs by the price per egg. She makes \$2 per fresh duck egg: 9 (fresh eggs left) × \$2 (price per egg) = \$18 daily revenue. Therefore, Janet makes \$18 per day at the farmers' market, selling the leftover fresh eggs from her ducks. & 236  & 0.5 & 0.29 & -0.29 &  0.84 \\
\textbf{(3) Partly Correct, Wrong Answer:} Janet's ducks lay 16 eggs per day. To determine how many fresh eggs she sells each day, we need to subtract the number of eggs she eats for breakfast and the number she uses for muffins: Fresh eggs daily = 16 eggs per day - 3 eggs for breakfast - 4 eggs for muffins = 16 - 3 - 4 = 2 eggs per day So, Janet sells 2 fresh duck eggs daily at the farmers' market. Since she sells each egg for \$2, her daily earnings can be calculated by multiplying the number of fresh ducks by their price: Daily earnings = Number of eggs sold x Price per egg = 2 eggs x \$2 per egg = \$4 Therefore, Janet makes \$4 every day at the farmers' market... (Truncated) & 457  & -0.5 & -1.5 & -0.83 & -0.16 \\
\textbf{(4) Irrelevant or Completely Wrong:} It is never 30 °C in winter. On the shortest day, in the Arctic Circle, the low is -17 °C. On the longest day, the low is 17 °C. Thus, 10 °C is the average low. We can transform this into a simpler version. Janet’s ducks lay 16 eggs per day, eat 3, use 4. Janet makes 13 eggs. She sells them for \$0.05 each: 13 × 0.05 = \$0.65... (Truncated) & 256 & -0.5 & -0.77 & -0.83 & -0.29 \\
\textbf{(5) Correct, but no reasoning (likely memorized):} The answer is 18. & 5 & -0.5 & 1.17 & 1.17 & -0.16 \\
\textbf{(6) Incorrect, and no reasoning:} The answer is 0. & 5  & -0.5 &  0.17 & 0.17 & -1.20 \\
\hline
\end{tabularx} }
\end{table}

\newpage
\subsection{Comparison of GRPO and SA-GRPO}
\label{app:case}

\subsubsection{Case 1: Improved Conciseness}

\begin{center}
\begin{tcolorbox}[casestyle]
\textbf{Question}: Let $r, s,$ and $t$ be solutions to the equation
\[
2x^3 - 3x^2 + 4x - 1 = 0.
\]
Find $r^2+s^2+t^2$.
\\\\
\textbf{Ground truth}: \boxed{-\frac{7}{4}}
\end{tcolorbox}
\end{center}
\textbf{Qwen3-1.7B + GRPO Answer}:
\\\\
{\small
To find \( r^2 + s^2 + t^2 \) where \( r, s, \) and \( t \) are the solutions to the equation \( 2x^3 - 3x^2 + 4x - 1 = 0 \), we can use the relationships between the roots and the coefficients of the polynomial. Specifically, we can use the identity:

\[
r^2 + s^2 + t^2 = (r + s + t)^2 - 2(rs + rt + st)
\]

First, we need to find the values of \( r + s + t \), \( rs + rt + st \), and \( rst \) using the coefficients of the polynomial.

The general form of a cubic polynomial is \( ax^3 + bx^2 + cx + d = 0 \). For our polynomial \( 2x^3 - 3x^2 + 4x - 1 = 0 \), we have:
\[
a = 2, \quad b = -3, \quad c = 4, \quad d = -1
\]

Using Vieta's formulas, we know:
\[
r + s + t = -\frac{b}{a} = -\frac{-3}{2} = \frac{3}{2},
\]
\[
rs + rt + st = \frac{c}{a} = \frac{4}{2} = 2,
\]
\[
rst = -\frac{d}{a} = -\frac{-1}{2} = \frac{1}{2}.
\]

Now, we can substitute these values into the identity for \( r^2 + s^2 + t^2 \):

\[
r^2 + s^2 + t^2 = (r + s + t)^2 - 2(rs + rt + st)
\]

Substituting the values we found:

\[
r^2 + s^2 + t^2 = \left( \frac{3}{2} \right)^2 - 2^2
\]

Calculating each term:

\[
\left( \frac{3}{2} \right)^2 = \frac{9}{4}, \qquad 2^2 = 4
\]

So,

\[
r^2 + s^2 + t^2 = \frac{9}{4} - 4
\]

To subtract these, we need a common denominator:

\[
4 = \frac{16}{4}
\]

Thus,

\[
r^2 + s^2 + t^2 = \frac{9}{4} - \frac{16}{4} = \frac{9 - 16}{4} = \frac{-7}{4}
\]

Therefore, the answer is:

\[
\boxed{-\frac{7}{4}}
\]}
\\\\
\textbf{Qwen3-1.7B + SA-GRPO Answer}:
\\\\
{\small
To solve this problem, we can use Vieta's formulas, which relate the coefficients of the polynomial to sums and products of its roots. For the polynomial 
\[
2x^3 - 3x^2 + 4x - 1 = 0,
\]
the roots are \(r, s,\) and \(t\).

Vieta's formulas tell us:
\begin{itemize}
  \item The sum of the roots \(r + s + t = -\frac{-3}{2} = \frac{3}{2}\).
  \item The sum of the product of the roots taken two at a time \(rs + rt + st = \frac{4}{2} = 2\).
  \item The product of the roots \(rst = -\frac{-1}{2} = \frac{1}{2}\).
\end{itemize}

We need to find \(r^2 + s^2 + t^2\). We can use the identity:
\[
r^2 + s^2 + t^2 = (r + s + t)^2 - 2(rs + rt + st).
\]

Substituting the values from Vieta's formulas:
\[
r^2 + s^2 + t^2 = \left(\frac{3}{2}\right)^2 - 2(2).
\]
\[
r^2 + s^2 + t^2 = \frac{9}{4} - 4.
\]
\[
r^2 + s^2 + t^2 = \frac{9}{4} - \frac{16}{4}.
\]
\[
r^2 + s^2 + t^2 = \frac{9 - 16}{4}.
\]
\[
r^2 + s^2 + t^2 = \frac{-7}{4}.
\]

Therefore, the answer is \(\boxed{-\frac{7}{4}}\).}

\subsubsection{Case 2: Improved Conciseness and Accuracy}

\begin{center}
\begin{tcolorbox}[casestyle]
\textbf{Question}: How many integers \(x\) satisfy
\[
|x| + 5 < 7 \quad \text{and} \quad |x-3| > 2 \, ?
\]

\textbf{Ground truth}: \boxed{2}
\end{tcolorbox}
\end{center}
\textbf{Qwen3-1.7B + GRPO Answer}:
\\\\
{\small
To solve the problem, we need to find the integers \( x \) that satisfy both inequalities:
\[
1.|x| + 5 < 7\ \ \ \ \ \ \ \ 
2. |x - 3| > 2
\]

Let's solve each inequality step by step.

Step 1: Solve the first inequality \( |x| + 5 < 7 \)

Subtract 5 from both sides:
\[
|x| < 2
\]

This absolute value inequality can be rewritten as:
\[
-2 < x < 2
\]

So, the integers \( x \) that satisfy this inequality are:
\[
x = -1, 0, 1
\]

Step 2: Solve the second inequality \( |x - 3| > 2 \)

This absolute value inequality can be split into two separate inequalities:
\[
x - 3 > 2 \quad \text{or} \quad x - 3 < -2
\]

Solving these inequalities:
\[
x > 5 \quad \text{or} \quad x < 1
\]

So, the integers \( x \) that satisfy this inequality are:
\[
x = \ldots, -2, -1, 0, 1, 2, 3, 4, 5, \ldots
\]

Step 3: Find the intersection of the solutions from both inequalities

We need to find the integers that satisfy both \( -2 < x < 2 \) and \( x > 5 \) or \( x < 1 \).

From the first inequality, the integers are:
\[
x = -1, 0, 1
\]

From the second inequality, the integers are:
\[
x = \ldots, -2, -1, 0, 1, 2, 3, 4, 5, \ldots
\]

The intersection of these two sets of integers is:
\[
x = -1, 0, 1
\]

Step 4: Count the number of integers in the intersection

The integers that satisfy both inequalities are:
\[
x = -1, 0, 1
\]

There are 3 integers in this set.

Therefore, the number of integers \( x \) that satisfy both inequalities is:
\[
\boxed{3}
\]
}
\\\\
\textbf{Qwen3-1.7B + SA-GRPO Answer}:
\\\\
{\small
To solve the problem, we need to analyze the two inequalities separately and then find the intersection of the solutions.

Step 1: Solve the first inequality \( |x| + 5 < 7 \)

Subtract 5 from both sides:
\[
|x| < 2
\]
This means that \( x \) must be within 2 units of 0:
\[
-2 < x < 2
\]

Step 2: Solve the second inequality \( |x - 3| > 2 \)

This inequality can be broken into two separate inequalities:
\[
x - 3 > 2 \quad \text{or} \quad x - 3 < -2
\]
Solving these, we get:
\[
x > 5 \quad \text{or} \quad x < 1
\]

Step 3: Find the intersection of the solutions

We need to find the values of \( x \) that satisfy both inequalities:
\[
-2 < x < 2
\]
\[
x > 5 \quad \text{or} \quad x < 1
\]

The intersection of these intervals is:
\[
-2 < x < 1
\]

Step 4: Identify the integers in the interval \( -2 < x < 1 \)

The integers in this interval are:
\[
-1, 0
\]

Step 5: Count the integers

There are 2 integers that satisfy both inequalities.

Therefore, the number of integers \( x \) that satisfy the given conditions is:
\[
\boxed{2}
\]
}

\end{document}